\documentclass[10pt,a4paper,logo]{vane_report}

\usepackage{amsmath,amssymb,mathtools}
\usepackage{graphicx,booktabs,array,multirow,tabularx}
\usepackage{colortbl}
\makeatletter
\let\VANEOriginalEndTabular\endtabular
\makeatother
\usepackage{arydshln}
\makeatletter
\let\endtabular\VANEOriginalEndTabular
\expandafter\let\csname endtabular*\endcsname\endtabular
\makeatother
\usepackage{wrapfig,etoolbox,xspace,placeins}
\usepackage{bm,makecell,subcaption}
\usepackage[authoryear,round]{natbib}
\usepackage{cleveref}
\usepackage{eso-pic}
\crefname{figure}{Figure}{Figures}
\Crefname{figure}{Figure}{Figures}
\crefname{table}{Table}{Tables}
\Crefname{table}{Table}{Tables}

\definecolor{groupbg}{RGB}{245,245,245}
\definecolor{napabg}{RGB}{237,232,248}
\definecolor{baselinebg}{RGB}{247,247,247}
\newcommand{\method}{VANE\xspace}

\newcommand{\miss}{\textemdash}
\newcommand{\blankfigure}[1]{%
  \begingroup\setlength{\fboxsep}{0pt}%
  \fbox{\parbox[c][#1][c]{\dimexpr\linewidth-2\fboxrule\relax}{\mbox{}}}%
  \endgroup}
\newcolumntype{Y}{>{\raggedright\arraybackslash}X}
\newcolumntype{Z}{>{\centering\arraybackslash}X}

\title{Learning from Next-Frame Prediction: Towards\\[3pt]
Unified and Scalable Visual World Modeling}

\makeatletter
\def\@author{%
  {\sffamily\bfseries\large
    Yifeng Yao$^{1,2}$,
    Jinghan Li$^{2}$,
    Zhihao Zhang$^{2,3}$,
    Yajiao Xiong$^{1,2}$,\\[3pt]
    Yang Jin$^{2,\ast}$,
    Kun Xu$^{2,\ast}$,
    Huishuai Zhang$^{1}$,
    Dongyan Zhao$^{1}$\par
  }
  \vspace{6pt}
  {\sffamily\normalsize
    $^{1}$Peking University\qquad
    $^{2}$ByteDance\qquad
    $^{3}$UNC-Chapel Hill
    \par
  }
}
\makeatother

\hypersetup{
  pdftitle={Learning from Next-Frame Prediction: Towards Unified and Scalable Visual World Modeling},
  pdfauthor={Yifeng Yao},
  pdfsubject={VANE}
}

\begin{abstract}
Video provides a natural source of supervision: what happens next serves as a learning signal for visual world modeling.
Yet masked video pretraining focuses on recovering missing content from surrounding context rather than predicting the future from the past. Autoregressive methods instead predict forward one token at a time, but have yet to match joint-embedding methods in representation quality.
This gap motivates us to revisit the granularity and space of future prediction. 
In this work, we introduce \textbf{\method{}}, a self-supervised visual pretraining framework built on next-frame prediction in embedding space.
A causal encoder summarizes past observations for a state predictor that infers the representation of the entire next frame, making temporal evolution the primary learning signal.
This prediction is carried out in a fixed embedding space provided by a frozen semantic tokenizer, emphasizing scene content and structure.
To retain fine-grained spatial detail alongside semantic content, a lightweight auxiliary generative decoder reconstructs future-frame patches conditioned on the predicted representations.
To unify visual modeling across images and videos, we construct a local-to-global crop sequence for still images, allowing the same encoder to predict newly revealed regions from earlier crops under the same objective.
Experiments show that \method{} improves with model and data scaling, achieving SOTA on most anticipation, recognition, and dense perception benchmarks among unified encoders.
Especially on EPIC-KITCHENS-100 action anticipation, it surpasses V-JEPA~2.1 by 7.1 points in mean-class recall@5.
These results demonstrate that next-frame prediction in embedding space is effective and scalable for visual world modeling.
\end{abstract}

\begin{document}
\maketitle

% Project leads: first-page footer, above the blue line
\AddToShipoutPictureFG*{%
  \AtPageLowerLeft{%
    \put(\LenToUnit{24mm},\LenToUnit{17mm}){%
      \sffamily\normalsize%
      $^{\ast}$Project leader%
    }%
  }%
}

% Start the introduction immediately after the abstract on page 1.
% =============================================================================
% 1. INTRODUCTION
% =============================================================================
\section{Introduction}
\label{sec:intro}

Humans come to understand the world by watching it unfold, mapping a continuous stream of sensory input into a stable sense of objects, motion, and events. What they acquire is not merely a catalogue of appearances, but an internal model of how the world evolves~\citep{rao1999predictive,lecun2022path}: what a scene contains, what is happening within it, and what is likely to happen next. Video captures the world in precisely this form, with temporal order providing a natural learning signal that requires no annotation. Predicting what comes next therefore requires a model to represent both the appearance and the dynamics of the visual world.

\begin{figure}[t]
  \centering
  \includegraphics[width=\linewidth]{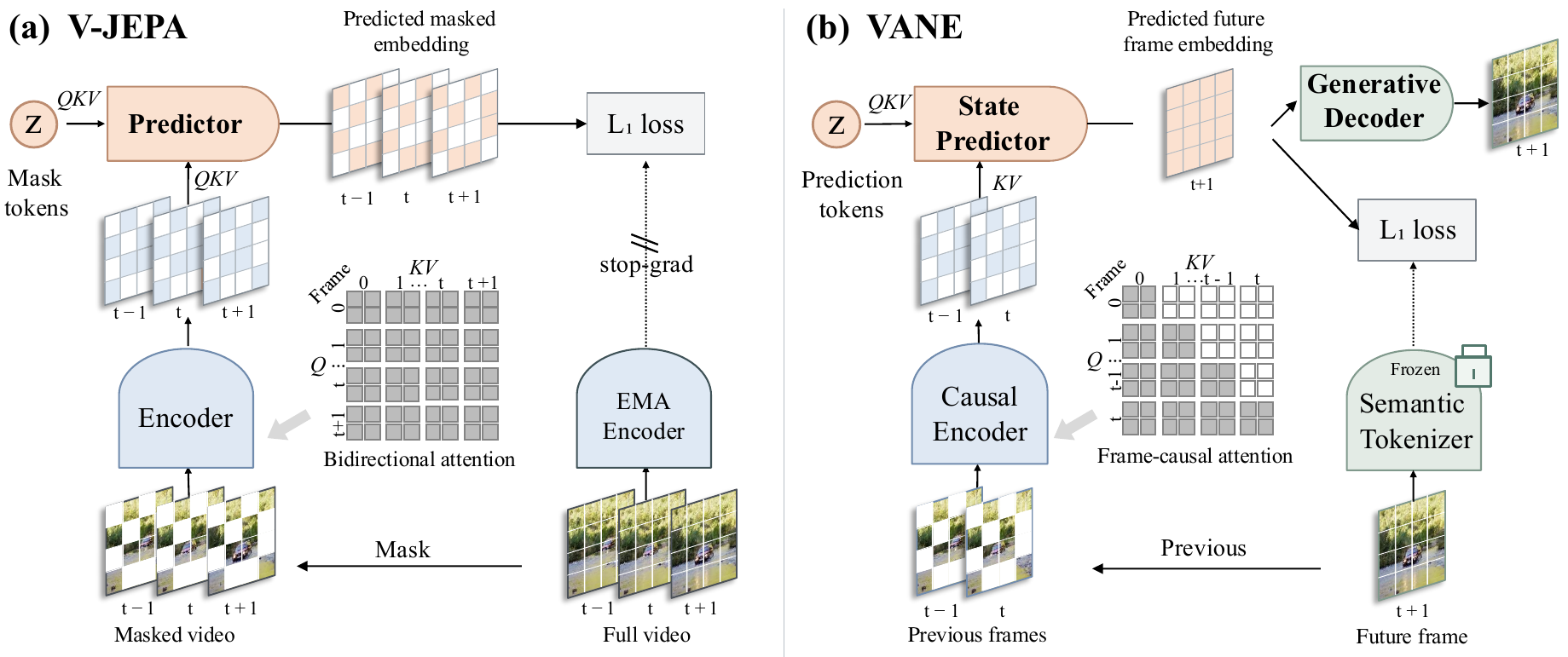}
  \caption{\textbf{From masked completion to next-frame prediction in
  embedding space.}
  (a)~V-JEPA masks part of a clip, encodes the visible patches with
  bidirectional attention, and predicts the embeddings of the masked
  regions, supervised by an EMA encoder of the full video.
  (b)~\method encodes previous frames with frame-causal attention and
  predicts the embedding of the entire next frame from them, with targets
  from a frozen semantic tokenizer. An auxiliary generative decoder,
  used only during pretraining, regularizes the predicted embedding to
  retain fine-grained spatial detail.}
  \label{fig:paradigm}
  \vspace{-4mm}
\end{figure}

Yet most video pretraining methods do not learn by predicting what comes next. Masked video models~\citep{feichtenhofer2022maest}, such as VideoMAE~\citep{tong2022videomae}, hide a large fraction of spatiotemporal patches and reconstruct their pixels from the visible ones~\citep{he2022mae}. Joint-embedding methods~\citep{assran2023ijepa}, such as V-JEPA~\citep{bardes2024vjepa,assran2025vjepa2}, likewise mask part of the video, but predict representations of the masked regions instead of their pixels (Fig.~\ref{fig:paradigm}a). Both remain forms of contextual recovery: the model infers what is missing from surrounding observations rather than predicting the future from the past. Autoregressive pretraining~\citep{chen2020igpt,elnouby2024aim}, extended to video by Toto~\citep{rajasegaran2025toto}, instead predicts forward in time by treating video as a sequence of visual tokens and learning through next-token prediction. While viable and scalable, this formulation has not matched joint-embedding methods in representation quality~\citep{rajasegaran2025toto}. This raises a central question: \emph{at what granularity, and in what space, should the future be predicted?}

In this work, we answer this question with \method, a self-supervised visual pretraining framework built on \textbf{next-frame prediction in embedding space}. Specifically, \method shifts prediction from individual tokens to the frame level and performs it in a semantic representation space. At the frame level, the distinction between spatial and temporal structure becomes explicit: structure within a frame is largely spatial, whereas changes across consecutive frames capture how objects, actions, and scenes evolve over time. Frame-level prediction therefore places temporal evolution at the center of the learning objective, rather than devoting much of the objective to completing individual tokens within the same frame. Additionally, prediction is carried out in semantic embedding space, building on the principle of joint-embedding learning~\citep{assran2023ijepa,bardes2024vjepa}. Semantic representations emphasize the higher-level content and structure of a scene, allowing the model to focus on the aspects of the visual world that persist and evolve over time.

% To achieve this objective, \method consists of four components (Fig.~\ref{fig:paradigm}b):
% (1) a causal encoder that processes observed frames in temporal order, with each frame attending only to itself and preceding frames, so that its representation summarizes how the scene has evolved up to that point;
% (2) a state predictor that uses this temporal context to predict the representation of the next frame;
% (3) a frozen semantic tokenizer, implemented as a pretrained image encoder~\citep{oquab2023dinov2}, that provides the prediction target. As in modern generative models operating in fixed latent spaces~\citep{rombach2022ldm}, freezing the tokenizer provides a stable target that cannot adapt to make the prediction task easier;
% and (4) a lightweight generative decoder that reconstructs patches of the future frame from the predicted embedding. This auxiliary reconstruction objective further regularizes the predicted embedding to preserve fine-grained spatial structure.

We instantiate this objective with four components
(Fig.~\ref{fig:paradigm}b).
A causal encoder summarizes the observed history through frame-causal attention, with each frame attending only to itself and preceding frames.
From this temporal context, a state predictor infers the representation of the next frame.
A frozen semantic tokenizer, implemented as a pretrained image
encoder~\citep{oquab2023dinov2}, provides a fixed semantic space in which future scene content and structure are predicted.
As in generative models operating in fixed latent spaces~\citep{rombach2022ldm}, we keep the tokenizer frozen to provide fixed prediction targets and support stable optimization.
Finally, a lightweight generative decoder reconstructs each future-frame patch conditioned on its predicted embedding.
This auxiliary reconstruction objective complements semantic prediction by encouraging the model to learn representations that preserve fine-grained spatial detail.

% Beyond video, \method naturally extends to still images under the same predictive objective. Images lack the temporal order that next-frame prediction relies on, so we construct one through \textbf{local-to-global observation}: an image is presented as a sequence of local-to-global crops, and the model predicts the representation of the next crop just as it predicts the next frame in video. This allows images and videos to be trained jointly from the start, yielding a unified visual encoder without separate pretraining stages or modality-specific objectives.
To unify visual modeling across images and videos, we extend the same predictive objective to still images. Unlike videos, still images lack a temporal order, so we introduce \textbf{local-to-global observation}: each image is presented as a sequence of local-to-global crops, and the model predicts representations of newly revealed regions in the next crop from preceding crops, just as it predicts the next frame from preceding frames in video. This enables joint image--video training, yielding a unified visual encoder without separate pretraining stages or modality-specific objectives.

% We evaluate VANE across predictive, dense, and global understanding tasks on both video and images, showing that a single pretrained model transfers effectively across a broad range of visual tasks and modalities. Its performance improves consistently as we scale both the amount of pretraining data and the model size. Across these evaluations, 
% % VANE consistently outperforms V-JEPA 2.1~\citep{mur2026vjepa21}, a state-of-the-art joint-embedding video encoder, despite using a strictly causal backbone, whereas V-JEPA 2.1 encodes the full clip bidirectionally. The improvement is particularly pronounced on anticipation: on EPIC-KITCHENS-100~\citep{damen2022epickitchens}, VANE raises mean-class recall@5 from 38.4 to \textbf{45.5} (+7.1). It also outperforms V-JEPA 2.1 on dense prediction and global recognition.
% despite using frame-causal rather than bidirectional encoding, \method{} improves over V-JEPA~2.1~\citep{mur2026vjepa21}, a state-of-the-art joint-embedding video encoder, on anticipation and across most global classification and dense prediction benchmarks. The improvement is particularly pronounced on anticipation: on
% EPIC-KITCHENS-100~\citep{damen2022epickitchens}, \method{} raises mean-class recall@5 from 38.4 to \textbf{45.5} (+7.1).

Extensive experiments demonstrate that \method{} unifies predictive and perceptual understanding within a single pretrained visual model. \method{} achieves state-of-the-art performance among unified image--video encoders on most evaluated benchmarks spanning anticipation, global recognition, and dense perception. On EPIC-KITCHENS-100~\citep{damen2022epickitchens}, \method{} achieves 45.5 action mean-class recall@5, improving over V-JEPA~2.1~\citep{mur2026vjepa21} by 7.1 points. Beyond its broad transferability, \method{} scales consistently with both model capacity and pretraining data. Together, these results establish next-frame prediction in embedding space as a unified and scalable pretraining paradigm for visual foundation models.

Our contributions are as follows:
\begin{itemize}[leftmargin=*, noitemsep, nolistsep]

\item We formulate visual pretraining as next-frame prediction in embedding space. Instead of completing masked content or reconstructing pixels, the model predicts the semantic embedding of the next frame from past frames, learning directly from how scenes evolve over time.

% We propose \method, which instantiates this objective with causal future prediction, a frozen semantic target, and auxiliary generative reconstruction, and extends it to still images through \textbf{local-to-global observation}, yielding a unified and scalable framework for image-video pretraining.
\item We instantiate this objective by combining a causal encoder--predictor, a semantic tokenizer, and auxiliary generative decoding, then extend it to images through local-to-global observation for unified and scalable image--video pretraining.

% \item We demonstrate that, despite using a strictly causal backbone, \method outperforms the bidirectional V-JEPA 2.1 across predictive, dense, and global understanding tasks, with the largest gain on anticipation (+7.1 mean-class recall@5 on EPIC-KITCHENS-100).
% \end{itemize}

\item We demonstrate that next-frame prediction in embedding space is an effective and scalable visual pretraining paradigm, achieving state-of-the-art performance among unified encoders across most anticipation, global recognition, and dense perception benchmarks.
\end{itemize}

% =============================================================================
% 2. RELATED WORK
% =============================================================================
\section{Related Work}
\label{sec:related}

\noindent\textbf{Masked visual pretraining.}
Masked visual pretraining learns representations by recovering content withheld from an observation.
Building on masked autoencoding~\citep{he2022mae}, VideoMAE~\citep{tong2022videomae,wang2023videomaev2} and MAE-ST~\citep{feichtenhofer2022maest} reconstruct masked spatiotemporal patches in pixel space, while the InternVideo series~\citep{wang2022internvideo,wang2024internvideo2,wang2025internvideo25} incorporates masked modeling into broader video pretraining frameworks.
The Joint-Embedding Predictive Architecture (JEPA)~\citep{lecun2022path} formulates prediction in representation space rather than pixel space.
I-JEPA~\citep{assran2023ijepa} instantiates this principle for images, while the V-JEPA family~\citep{bardes2024vjepa,assran2025vjepa2,mur2026vjepa21} extends it to video, scaling masked feature prediction and improving global and dense representations.
Related methods distill pretrained visual features~\citep{wang2023mvd,li2023unmaskedteacher}, and Rethinking JEPA~\citep{li2026rethinkingjepa} shows that a frozen target encoder can support effective masked latent prediction.
InternVideo-Next~\citep{wang2026internvideonext} combines semantic alignment, generative reconstruction, and fixed-target feature prediction within a two-stage masked modeling framework.
Despite differences in target construction and supervision, these pretraining objectives remain centered on recovering or matching content within the observed clip rather than explicitly predicting its temporal continuation.
By predicting future-frame embeddings from past observations, \method{} makes temporal evolution central to pretraining while preserving fine-grained spatial structure through auxiliary reconstruction.

\noindent\textbf{Autoregressive and predictive visual learning.}
Autoregressive visual pretraining depends on how visual observations are represented and ordered.
iGPT~\citep{chen2020igpt} and AIM~\citep{elnouby2024aim} predict pixels or patches, while Toto~\citep{rajasegaran2025toto} learns from images and videos through next-token prediction over discrete visual codes.
Alternative formulations vary the prediction order, granularity, or target space: ARVideo~\citep{ren2024arvideo} randomizes spatiotemporal ordering, VAR~\citep{tian2024var} predicts progressively finer-resolution token maps, and NEPA~\citep{xu2025nepa} predicts successive image patch embeddings with stop-gradient targets.
A complementary line learns from future visual features through regression~\citep{vondrick2016anticipating} or contrastive objectives~\citep{oord2018cpc,han2019dpc}.
More recently, DINO-Foresight~\citep{karypidis2025foresight} and DINO-world~\citep{baldassarre2025dinoworld} forecast future features for scene understanding, while DINO-WM~\citep{zhou2025dinowm} uses action-conditioned feature prediction for planning.
Causal-JEPA~\citep{causaljepa2026} explores object-level latent masking to learn interaction-aware dynamics.
These recent world models learn prediction on top of pretrained visual encoders that remain fixed during dynamics learning.
\method{} instead uses next-frame embedding prediction to train the causal encoder itself, extending the same objective to unified image--video pretraining through local-to-global image views.
\section{Method}
\label{sec:method}
\label{sec:method_overview}

% We build \method{} around next-frame prediction in embedding space
% for unified and scalable visual pretraining; Fig.~\ref{fig:method} gives an
% overview. We first formulate next-frame prediction for videos and extend the same objective to still images through local-to-global next-crop prediction (Sec.~\ref{sec:ordered_prediction}). A \ourencoder summarizes
% the observed history, from which a context-isolated \ourpredictor predicts a
% state for each target patch (Sec.~\ref{sec:causal_predictor}). A frozen
% \ourtokenizer provides the semantic prediction space, while an auxiliary
% patch-wise generative decoder regularizes the predicted states to retain
% fine-grained spatial information
% (Secs.~\ref{sec:semantic_targets}--\ref{sec:generative_decoder}). Finally, we
% combine the semantic and generative objectives for unified image--video
% pretraining (Sec.~\ref{sec:objective}).

\begin{figure*}[t]
    \centering
    \IfFileExists{figures/method.pdf}{%
        \includegraphics[width=\textwidth]{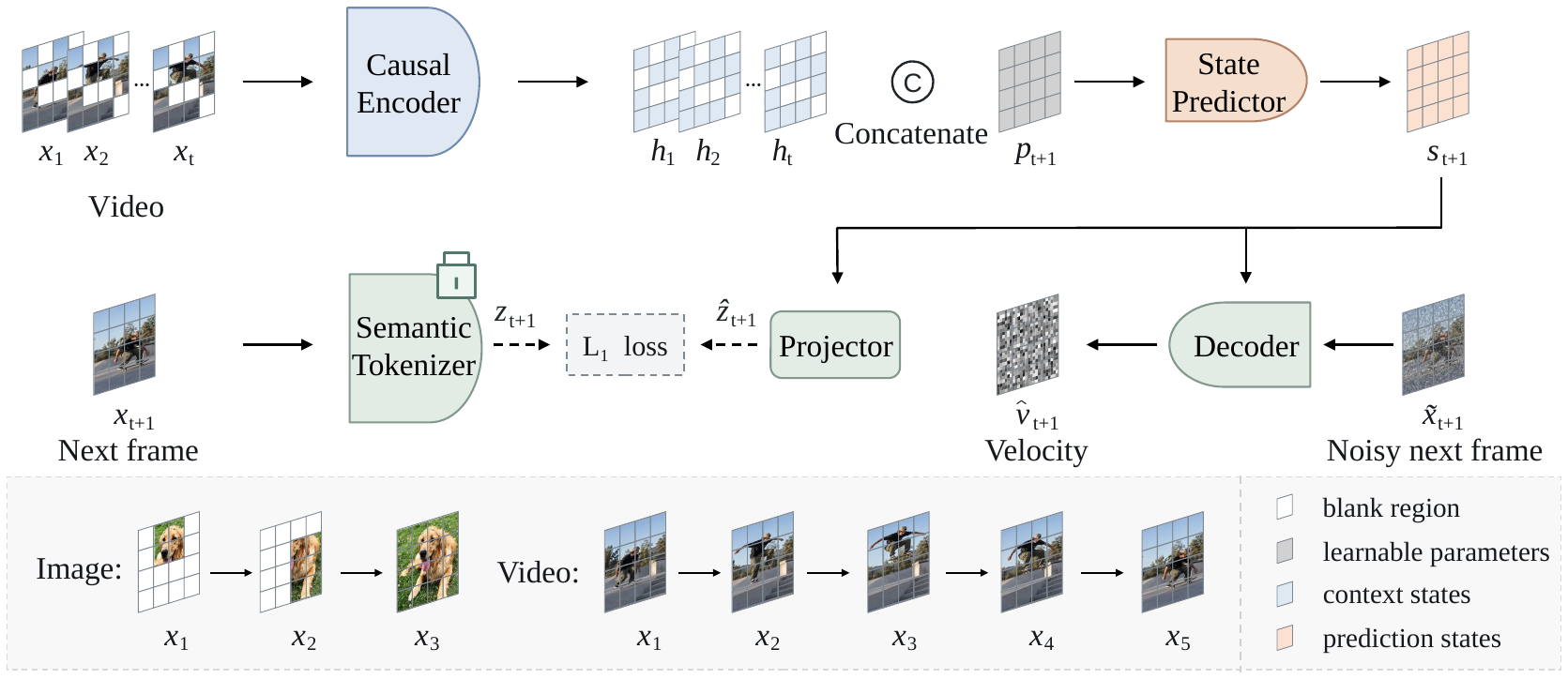}%
    }{\blankfigure{1.65in}}
    \caption{\textbf{Overview of \method.}
    \method{} predicts the embedding of the next frame from the frames
    observed so far.
    \emph{Top:} the causal encoder maps the observed frames
    $x_1,\ldots,x_t$ to context states $h_1,\ldots,h_t$, which are
    concatenated with prediction tokens $p_{t+1}$; the state predictor uses
    the context states only as keys and values and outputs the predicted
    state $s_{t+1}$.
    \emph{Middle:} a projector maps $s_{t+1}$ to the embedding space of the
    frozen semantic tokenizer, where it is supervised by the embedding of
    the next frame $x_{t+1}$ with an $L_1$ loss; the generative decoder,
    conditioned on $s_{t+1}$, predicts the flow velocity $\hat v_{t+1}$ from
    the noisy next frame $\tilde x_{t+1}$.
    \emph{Bottom:} images are presented as local-to-global crops and videos
    as frame sequences, so both share the same objective.}
    \label{fig:method}
    \vspace{-4mm}
\end{figure*}

This section presents \method{} (Fig.~\ref{fig:method}). We first
formulate next-frame prediction in embedding space
(Sec.~\ref{sec:semantic_targets}). Realizing this objective raises three
questions, which we address in turn: how to encode the observed frames
without access to the future (Sec.~\ref{sec:causal_encoder}), how to
predict the next frame from this encoding (Sec.~\ref{sec:causal_predictor}),
and how to retain the fine-grained spatial detail that semantic targets
abstract away (Sec.~\ref{sec:generative_decoder}). Finally, we extend the
objective to still images, which lack temporal order, for unified
image--video pretraining (Sec.~\ref{sec:image_crops}).

\subsection{Next-Frame Prediction in Embedding Space}
\label{sec:semantic_targets}
\label{sec:ordered_prediction}

To make temporal evolution the primary learning signal, \method{} learns
by predicting what comes next. Given a video clip
$(x_1,\ldots,x_N)$ of $N$ frames, the model predicts each frame
$x_{t+1}$ from the frames observed so far, $x_{\le t}$. Every transition
in the clip thus provides supervision, and each intermediate frame serves
both as a target given its predecessors and as context for its
successors.

Rather than predicting pixels, which would require modeling low-level
appearance details, we predict the next frame in a semantic embedding
space. This space is defined by a frozen semantic tokenizer $F$, a
pretrained image encoder that maps each frame to embeddings representing
its content and structure while abstracting over appearance variation. We keep $F$ fixed throughout pretraining so that the
target space cannot co-adapt with the model to simplify the prediction
task, as in generative models trained in fixed latent
spaces~\citep{rombach2022ldm}. The objective is thus
\begin{equation}
\min_{f}\;\sum_{t=1}^{N-1} d\Bigl(\underbrace{f\bigl(x_{\le t}\bigr)}_{\text{prediction}},\;\underbrace{F\bigl(x_{t+1}\bigr)}_{\text{target}}\Bigr),
\label{eq:objective}
\end{equation}
where $f$ is the predictive model and $d$ is a distance in the embedding
space (instantiated in Sec.~\ref{sec:causal_predictor}).

% This differs from feature distillation, in which a student
% regresses a teacher's features of the same input. Here, the model never
% observes the target frame $x_{t+1}$: $F$ only specifies what to predict,
% and the representation needed to predict it must be learned from the past
% frames alone.

\subsection{Causal Encoder}
\label{sec:causal_encoder}

We realize $f$ with a causal encoder followed by a state predictor. To
supervise every transition in a clip within a single forward pass, the
encoder must represent each frame using only the frames observed so far;
otherwise the target of a transition would leak into its own context.
The encoder $E_\theta$ therefore processes the context frames
$x_1,\ldots,x_{N-1}$ with \emph{frame-causal} attention
(Fig.~\ref{fig:paradigm}b): attention is unidirectional
across frames, so that tokens of frame $t$ never attend to later frames,
and bidirectional within each frame, since the patches of a frame are
observed at the same time and have no natural order. The resulting
context state $h_t$ thus depends only on frames up to $t$:
\begin{equation}
h_t=E_\theta\bigl(x_{\le t}\bigr),
\qquad 1\leq t<N.
\label{eq:causal_encoding}
\end{equation}
We apply tube masking~\citep{tong2022videomae,bardes2024vjepa} to the
context frames, so that $E_\theta$ receives only their visible patches,
while the prediction targets remain unmasked
(Appendix~\ref{app:video_masking}).

% \begin{figure}[t]
%     \centering
%     \includegraphics[width=0.6\linewidth]{figures/attention_mask.pdf}
%     \caption{\textbf{Attention masks.} \emph{Left:} frame-causal attention
%     in the causal encoder, bidirectional within each frame and
%     unidirectional across frames. \emph{Right:} context-isolated causal
%     attention in the state predictor. Prediction tokens for frame $t+1$ read
%     the context states of frames up to $t$, which supply only keys and
%     values, and attend to each other only within the same target frame.}
%     \label{fig:attention_mask}
% \end{figure}

\subsection{State Predictor and Semantic Objective}
\label{sec:causal_predictor}

\begin{wrapfigure}{r}{0.42\textwidth}
\vspace{-12pt}
\centering
\includegraphics[width=\linewidth]{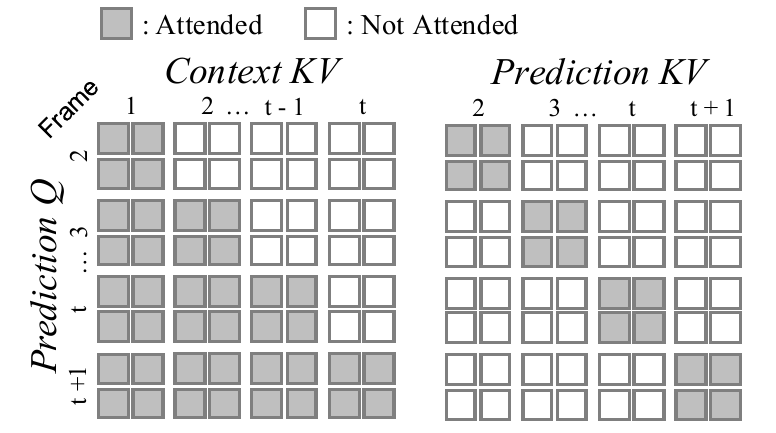}
\vspace{-14pt}
\caption{\textbf{State predictor attention.} Context states supply only
keys and values; prediction tokens attend within their own target frame.}
\label{fig:predictor_attention}
\vspace{-4pt}
\end{wrapfigure}
The state predictor $P_\phi$ then infers the representation of the next
frame from the encoded history. It places copies of a learned token $p$
at the positions of frame $t+1$, forming prediction tokens $p_{t+1}$. As
in the encoder, the prediction tokens of all target frames are processed
together in a single forward pass, under two attention constraints
(Fig.~\ref{fig:predictor_attention}). Under \emph{context isolation},
context states supply only keys and values and are not updated inside
the predictor, so that the encoder, rather than the predictor, must carry
the historical information needed for prediction. Under \emph{target
isolation}, prediction tokens for frame $t+1$ attend only to
$h_{\le t}=(h_1,\ldots,h_t)$ and to one another, so that each frame is
predicted jointly over its spatial extent from its own history alone,
independently of the other targets:
\begin{equation}
s_{t+1}=P_\phi\bigl(p_{t+1};h_{\le t}\bigr),
\qquad 1\leq t<N,
\label{eq:prediction}
\end{equation}
where $s_{t+1}$ is the predicted state of frame $t+1$.
To supervise the prediction in the semantic embedding space, a learned
projection $g_\omega$ maps the predicted state to the tokenizer space,
$\hat z_{t+1}=g_\omega(s_{t+1})$, and we minimize its distance to the
frozen target $z_{t+1}=F(x_{t+1})$ with an $L_1$ loss:
\begin{equation}
\mathcal L_{\mathrm{sem}}=\bigl\|\hat z_{t+1,j}-z_{t+1,j}\bigr\|_1,
\label{eq:semantic_loss}
\end{equation}
where $\hat z_{t+1,j}$ and $z_{t+1,j}$ denote the prediction and target at
patch $j$ of frame $t+1$, averaged over all transitions $t$ and supervised
patches $j$.
The encoder, predictor, and projection thus together form the predictive
model of Eq.~\eqref{eq:objective}:
$f(x_{\le t})=g_\omega\bigl(P_\phi(p_{t+1};h_{\le t})\bigr)$, with $h_{\le t}$
given by Eq.~\eqref{eq:causal_encoding}.
% For videos, the
% whole next frame is supervised; for images, only newly revealed regions
% are (Sec.~\ref{sec:image_crops}).

\subsection{Auxiliary Generative Reconstruction}
\label{sec:generative_decoder}

Semantic prediction emphasizes abstract visual content but may
underrepresent fine-grained spatial information useful for dense
perception. We therefore introduce a lightweight auxiliary generative decoder
$D_\psi$ that regularizes each predicted state through patch-level
reconstruction. Let $x_{t+1,j}$ and $s_{t+1,j}$ denote the pixels and
predicted state of patch $j$ of frame $t+1$. The decoder reconstructs
each supervised patch conditioned on $s_{t+1,j}$, and the reconstruction
loss is back-propagated through the predictor into the encoder.

Since the next frame is not fully determined by the past, a
deterministic pixel regressor would be driven toward the average of
plausible outcomes. We therefore model the distribution of each patch
given its predicted state, as in per-token diffusion
heads~\citep{li2024mar}, and formulate the decoder as a conditional flow
matching model~\citep{lipman2023flow}. Each supervised patch is
corrupted as $\tilde x_{t+1,j}=(1-\tau)x_{t+1,j}+\tau\epsilon_{t+1,j}$
with $\epsilon_{t+1,j}\sim\mathcal N(0,I)$ and $\tau\sim\mathcal U[0,1]$,
and the decoder predicts the velocity $\epsilon_{t+1,j}-x_{t+1,j}$ with
an $L_1$ loss:
\begin{equation}
\mathcal L_{\mathrm{flow}}
=\mathbb E_{\tau,\epsilon}\bigl\|D_\psi\bigl(\tilde x_{t+1,j},\tau,s_{t+1,j}\bigr)
-(\epsilon_{t+1,j}-x_{t+1,j})\bigr\|_1,
\label{eq:flow_loss}
\end{equation}
averaged over the same transitions and supervised patches as
$\mathcal L_{\mathrm{sem}}$.

\subsection{Unified Image--Video Pretraining}
\label{sec:image_crops}

To train a unified encoder on both images and videos with the same
objective, we arrange each still image into an ordered sequence of views
through local-to-global observation. The sequence consists of two local
crops $x_1$ and $x_2$ of increasing area followed by a larger global crop
$x_3$ ($K=3$; Appendix~\ref{app:image_sampling}). Like successive frames,
successive crops reveal new content that must be inferred from what has
been observed so far (examples in
Appendix~Fig.~\ref{fig:next_crop_visualization}). Each crop is treated as
one frame by the encoder and predictor, and all crops are aligned to the
patch grid of the full image. Let $c_t$ be the binary mask of the region
covered by crop $t$ on this grid. Supervision is restricted to the newly
revealed region, i.e., the part of crop $t+1$ covered by none of the
preceding crops:
\begin{equation}
m_{t+1}=c_{t+1}\odot\prod_{i=1}^{t}\bigl(1-c_i\bigr),
\label{eq:image_targets}
\end{equation}
where $\odot$ denotes element-wise multiplication, and both losses are
averaged over the positions where $m_{t+1}=1$. Regions covered by earlier
crops are excluded because they are already visible in the context, so
predicting them would amount to reconstructing the input rather than
inferring unseen content. This progressively expanding context transfers
better than both a fixed order over image cells and masked-image
completion (Table~\ref{tab:ablations}b).
% The tokenizer encodes the full image once to provide the targets,
% while the encoder and predictor never see it. 

\noindent\textbf{Training objective.}
\label{sec:objective}
We jointly pretrain on images and videos by combining semantic
prediction and generative reconstruction:
\begin{equation}
\mathcal L=\mathcal L_{\mathrm{sem}}+\lambda_{\mathrm{flow}}\,\mathcal L_{\mathrm{flow}},
\label{eq:total_loss}
\end{equation}
where $\lambda_{\mathrm{flow}}$ weights the reconstruction loss. Both
losses are computed separately on video and image batches and averaged
with equal weight, and all components except the frozen tokenizer are
optimized jointly (Appendix~\ref{app:training}).

% Layout requirements: add these packages to the main document preamble.
% \usepackage{capt-of}
% \usepackage{wrapfig}
% Existing manuscript packages/macros (tabularx, Y, booktabs, row colors, etc.) are retained.

\section{Experiments}
\label{sec:exp}

We evaluate whether next-frame prediction provides an effective objective for visual world modeling on anticipation, global classification, and dense perception tasks, and then ablate the design choices.

\noindent\textbf{Implementation details.}
Our main model pairs a 1B-parameter ViT-g encoder with a frozen DINOv2
ViT-L~\citep{oquab2023dinov2} semantic tokenizer. It is pretrained on
\textsc{VISTA}, our curated image--video corpus, which combines public
image datasets (ImageNet-21K~\citep{deng2009imagenet},
SA-1B~\citep{kirillov2023sam}) and video datasets
(SSv2~\citep{goyal2017something},
K710~\citep{kay2017kinetics,li2023uniformerv2},
HowTo100M~\citep{miech2019howto100m}) with additional retrieved and
class-balanced data. Training proceeds in four
stages: warm-up and base pretraining on 16-frame clips at
$256\times256$, long-video post-training on 64-frame clips, and a
high-resolution cooldown at $384\times384$. The first two stages take
about 26.6k B200 GPU-hours and the last two a further 30.7k.
% (Appendices~\ref{app:architecture} and~\ref{app:training}).

\noindent\textbf{Evaluation protocol.}
All evaluations use frozen features: the pretrained encoder is kept
fixed, and a lightweight probe is trained on its features for each task.
Probes follow the protocols of V-JEPA~2.1~\citep{mur2026vjepa21} unless
otherwise stated.
% The predictor
% and generative decoder are not used. 
We compare primarily with V-JEPA~2 and V-JEPA~2.1 at
the same 1B scale; other encoders are either taken from the literature
or re-evaluated under the same protocol, as marked in each table.

\begin{table}[t]
\centering
\small
\setlength{\tabcolsep}{5pt}
\caption{\textbf{Action anticipation on EPIC-KITCHENS-100.}
Mean-class recall@5 (\%) for verb, noun, and action on the validation set,
with a one-second anticipation gap. Bold denotes the best result.}
\label{tab:ek100}
\tabcolsep=0.18cm
\resizebox{0.7\linewidth}{!}{
\begin{tabular}{lcccc}
\toprule
Method & Param. & Verb & Noun & Action \\
\midrule
InAViT~\citep{roy2024inavit} & 160M & 51.9 & 52.0 & 25.8 \\
Video-LLaMA~\citep{zhang2023videollama} & 7B & 52.9 & 52.0 & 26.0 \\
PlausiVL~\citep{mittal2024plausivl} & 8B & 55.6 & 54.2 & 27.6 \\
\midrule
V-JEPA 2 ViT-g~\citep{assran2025vjepa2} & 1B & 63.6 & 57.1 & 39.7 \\
V-JEPA 2.1 ViT-g~\citep{mur2026vjepa21} & 1B & 63.6 & 56.2 & 38.4 \\
\rowcolor{napabg}
\method{} (ours) & 1B & \textbf{65.7} & \textbf{67.6} & \textbf{45.5} \\
% V-JEPA 2.1 ViT-G~\citep{mur2026vjepa21} & 2B & 64.3 & 59.9 & 40.8 \\
\bottomrule
\end{tabular}}
\vspace{-4mm}
\end{table}

\subsection{Anticipation}
\label{sec:future}

We first examine whether next-frame pretraining supports anticipation,
on action anticipation (EK100) and short-term object-interaction
anticipation (Ego4D).

\noindent\textbf{Action anticipation.}
On EPIC-KITCHENS-100 (EK100)~\citep{damen2022epickitchens}, we follow the
V-JEPA~2.1~\citep{mur2026vjepa21} protocol with a one-second anticipation
gap and report mean-class recall@5 on the validation set. The attentive
probe uses only frozen encoder features; the state predictor is not used.
\method{} reaches 45.5 on actions, exceeding the 1B V-JEPA~2.1 and
V-JEPA~2 by 7.1 and 5.8 points, respectively (Table~\ref{tab:ek100}),
with the largest gain on nouns (67.6 vs.\ 56.2). It also outperforms the
8B PlausiVL by 17.9 points.
% The gains in action anticipation suggest that next-frame prediction in embedding space helps the model understand how visual events evolve and what is likely to happen next, a central goal of visual world modeling.
These gains suggest that next-frame
prediction in embedding space encourages representations that capture
how visual events evolve and what is likely to happen next.

\begin{table*}[t]
\centering
\vspace{4mm}
\scriptsize
\setlength{\tabcolsep}{2.2pt}
\caption{\textbf{Short-term object-interaction anticipation on Ego4D STA v2.}
The model localizes the next-active object and predicts its noun, the
interaction verb, and the time to contact. Each column reports Top-5 AP or
mAP counting a prediction as correct only if all listed components are
correct: box ($b$), noun ($N$), verb ($V$), and time to contact
($\delta$). Bold denotes the best results.}
\label{tab:ego4d}
\resizebox{\textwidth}{!}{%
\begin{tabular}{lcccccccccccc}
\toprule
Model
& \shortstack{AP\\$b$}
& \shortstack{AP\\$b{+}N$}
& \shortstack{AP\\$b{+}V$}
& \shortstack{AP\\$b{+}\delta$}
& \shortstack{AP\\$b{+}N{+}V$}
& \shortstack{AP\\$b{+}\delta{+}N$}
& \shortstack{AP\\$b{+}\delta{+}V$}
& \shortstack{AP\\All}
& \shortstack{mAP\\$N$}
& \shortstack{mAP\\$N{+}V$}
& \shortstack{mAP\\$N{+}\delta$}
& \shortstack{mAP\\All} \\
\midrule
\multicolumn{13}{l}{\textit{Results reported in the literature}}\\
StillFast~\citep{ragusa2023stillfast} & \miss & \miss & \miss & \miss & \miss & \miss & \miss & \miss & 20.2 & 10.3 & 7.17 & 3.96\\
STAformer~\citep{murlabadia2024staformer} & 38.3 & 28.3 & 16.6 & 12.3 & 12.8 & 8.89 & 5.47 & 4.06 & 29.4 & 15.3 & 9.94 & 5.67\\
% GANO~\citep{thakur2023guidedattention} & 45.3 & 27.0 & 12.2 & 16.6 & 6.54 & 9.0 & 4.18 & 2.47 & \miss & \miss & \miss & \miss\\
\midrule
\multicolumn{13}{l}{\textit{Evaluated using the same protocol}}\\
% DINOv2 ViT-L~\citep{oquab2023dinov2} & 45.1 & 37.8 & 21.9 & 14.4 & 17.6 & 10.6 & 7.27 & 5.25 & 28.3 & 15.2 & 9.22 & 5.25\\
DINOv2 ViT-g~\citep{oquab2023dinov2} & 47.8 & 38.1 & 23.6 & 14.1 & 19.1 & 10.9 & 7.29 & 5.73 & 30.6 & 16.8 & 9.37 & 5.44\\
DINOv3 ViT-H+~\citep{simeoni2025dinov3} & 48.9 & 40.2 & 23.4 & 13.6 & 19.4 & 10.8 & 6.71 & 5.45 & 32.4 & 17.2 & 9.53 & 5.20\\
DINOv3 ViT-7B~\citep{simeoni2025dinov3} & 50.2 & 40.9 & 23.8 & 14.2 & 19.8 & 11.5 & 7.33 & 5.82 & 33.8 & 17.4 & 10.3 & 5.68\\
V-JEPA 2 ViT-g~\citep{assran2025vjepa2} & 45.7 & 34.4 & 22.2 & 16.7 & 16.9 & 12.1 & 8.56 & 6.26 & 27.2 & 14.9 & 10.4 & 6.02\\
V-JEPA 2.1 ViT-g~\citep{mur2026vjepa21} & 47.3 & 36.5 & 23.6 & 18.1 & 18.2 & 13.5 & 9.55 & 7.21 & 28.7 & 15.5 & 11.4 & 6.75\\
% \midrule
\rowcolor{napabg}
\method{} ViT-g & \textbf{57.9} & \textbf{46.4} & \textbf{24.6} & \textbf{21.8} & \textbf{19.9} & \textbf{17.1} & \textbf{10.2} & \textbf{8.18} & \textbf{37.6} & \textbf{17.7} & \textbf{14.7} & \textbf{7.81} \\
% V-JEPA 2.1 ViT-G (2B) & 50.7 & 39.3 & 25.8 & 20.2 & 20.3 & 15.1 & 10.8 & 8.20 & 30.9 & 17.2 & 12.8 & 7.71\\
\bottomrule
\end{tabular}}
\vspace{-4mm}
\end{table*}

\begin{table*}[t]
    \centering
    \small
    \caption{\textbf{Global classification.}
Top-1 accuracy (\%) of attentive probes on frozen features, covering
appearance understanding (IN1K, K400) and motion understanding (SSv2,
Diving-48). Avg.\ is reported for models evaluated on all four
tasks. Bold denotes the best result in image--video encoders.}
    \label{tab:global_tasks}
    \resizebox{\textwidth}{!}{%
    \begin{tabular}{lcccccc}
        \toprule
        \textbf{Method}
        & \textbf{Param.}
        & \multicolumn{2}{c}{\textbf{Appearance Understanding}}
        & \multicolumn{2}{c}{\textbf{Motion Understanding}}
        & \textbf{Avg.} \\
        \cmidrule(lr){3-4}\cmidrule(lr){5-6}
        & & IN1K & K400 & SSv2 & Diving-48 & \\
        \midrule
        \multicolumn{7}{l}{\textit{Results reported in the literature}} \\
        Toto~\citep{rajasegaran2025toto}              & 1B  & 75.3   & 74.4   & --   & --   & -- \\
        VideoMAEv2~\citep{wang2023videomaev2}        & 1B  & 71.4 & 82.8 & 56.1 & --   & -- \\
        InternVideo2-1B~\citep{wang2024internvideo2}   & 1B  & --   & 87.9 & 67.3 & --   & -- \\
        VideoPrism~\citep{zhao2024videoprism}        & 1B  & --   & 87.6 & 68.5 & 71.3 & -- \\
        % InternVideo2$_{\mathrm{2S}}$-1B~\citep{wang2024internvideo2} & 1B & 85.8 & 89.4 & 69.7 & 86.4 & 82.8 \\
        DINOv3~\citep{simeoni2025dinov3}            & 7B  & 88.4 & 87.8 & 70.1 & --   & -- \\
        \midrule
        \multicolumn{7}{l}{\textit{Image encoders evaluated under the same protocol}} \\
        DINOv2~\citep{oquab2023dinov2}            & 1.1B & 86.1 & 83.6 & 50.7 & 82.5 & 75.7 \\
        DINOv3 ViT-H+~\citep{simeoni2025dinov3}     & 0.8B & 88.2 & 86.6 & 63.1 & 85.3 & 80.8 \\
        $\mathrm{PE}_{\mathrm{core}}$ G~\citep{bolya2025perceptionencoder} & 1.9B & 87.6 & 88.5 & 55.4 & 76.9 & 77.1 \\
        SigLIP 2~\citep{tschannen2025siglip2}           & 1.2B & 88.0 & 87.3 & 49.9 & 75.3 & 75.1 \\
        \midrule
        \rowcolor{baselinebg}
        \multicolumn{7}{l}{\textit{Image--video encoders evaluated under the same protocol}} \\
        \rowcolor{baselinebg}
        V-JEPA~\citep{bardes2024vjepa}            & 630M & 75.9 & 82.0 & 71.4 & 87.9 & 79.3 \\
        \rowcolor{baselinebg}
        V-JEPA 2~\citep{assran2025vjepa2}          & 1B   & 84.6 & 86.5 & 75.5 & \textbf{90.1} & 84.2 \\
        \rowcolor{baselinebg}
        V-JEPA 2.1~\citep{mur2026vjepa21}        & 1B   & 83.7 & 86.2 & 75.3 & 89.0 & 83.6 \\
        \rowcolor{napabg}
        \method{} (ours)         & 1B   & \textbf{87.5} & \textbf{88.6} & \textbf{75.9} & 89.7 & \textbf{85.4} \\
        \bottomrule
    \end{tabular}}
% \vspace{2pt}
\vspace{-4mm}
% \parbox{\textwidth}{\scriptsize $\dagger$ Evaluation setting to be confirmed.}
\end{table*}

\noindent\textbf{Short-term object-interaction anticipation.}
Ego4D STA v2~\citep{grauman2022ego4d} additionally requires localizing the
next-active object and predicting its category, the interaction verb,
and the time to contact. Following V-JEPA~2.1, we train an attentive
probe with frame-guided temporal pooling and a detection head on frozen
features and report Top-5 AP and mAP. \method{} reaches 7.81 mAP All,
compared with 6.75 for V-JEPA~2.1 and 6.02 for V-JEPA~2, and improves box
AP by 10.6 points over V-JEPA~2.1 (Table~\ref{tab:ego4d}). It outperforms
all encoders evaluated under the same protocol, including DINOv3 ViT-7B,
on all twelve metrics. Together with EK100, these results suggest that
\method{} anticipates not only what happens next but also where and when
it happens. Qualitative examples are provided in
Appendix~\ref{app:ego4d_visualization}.

\subsection{Global Classification}
\label{sec:global}

To assess recognition beyond anticipation, we evaluate image
classification on IN1K~\citep{deng2009imagenet} and action recognition on
K400, SSv2, and Diving-48~\citep{kay2017kinetics,goyal2017something,li2018diving48}.
Following V-JEPA~2.1, we train four-layer attentive probes on frozen
features with $256\times256$ inputs and report top-1 accuracy. Under the
same protocol, \method{} outperforms V-JEPA~2.1 on all four benchmarks,
by 3.8 points on IN1K, 2.4 on K400, 0.6 on SSv2, and 0.7 on Diving-48,
raising the average from 83.6 to 85.4 (Table~\ref{tab:global_tasks});
V-JEPA~2 remains slightly better on Diving-48 (90.1 vs.\ 89.7).
\method{} also exceeds the published results of Toto, which
autoregressively predicts future visual tokens, by 12.2 points on IN1K
and 14.2 on K400. Against image encoders, \method{} nearly matches DINOv3
ViT-H+ on IN1K (87.5 vs.\ 88.2) while exceeding it by 12.8 points on SSv2
(75.9 vs.\ 63.1), suggesting that unified image--video pretraining
retains appearance understanding while adding motion understanding.
Notably, it also surpasses DINOv2 ViT-g, a larger variant of its own
frozen tokenizer, by 25.2 points on SSv2 (75.9 vs.\ 50.7), indicating
that next-frame prediction learns temporal structure absent from its
targets.

\subsection{Dense Perception}
\label{sec:dense}

\begin{wrapfigure}{r}{0.45\textwidth}
\vspace{-12pt}
\centering
\includegraphics[width=\linewidth]{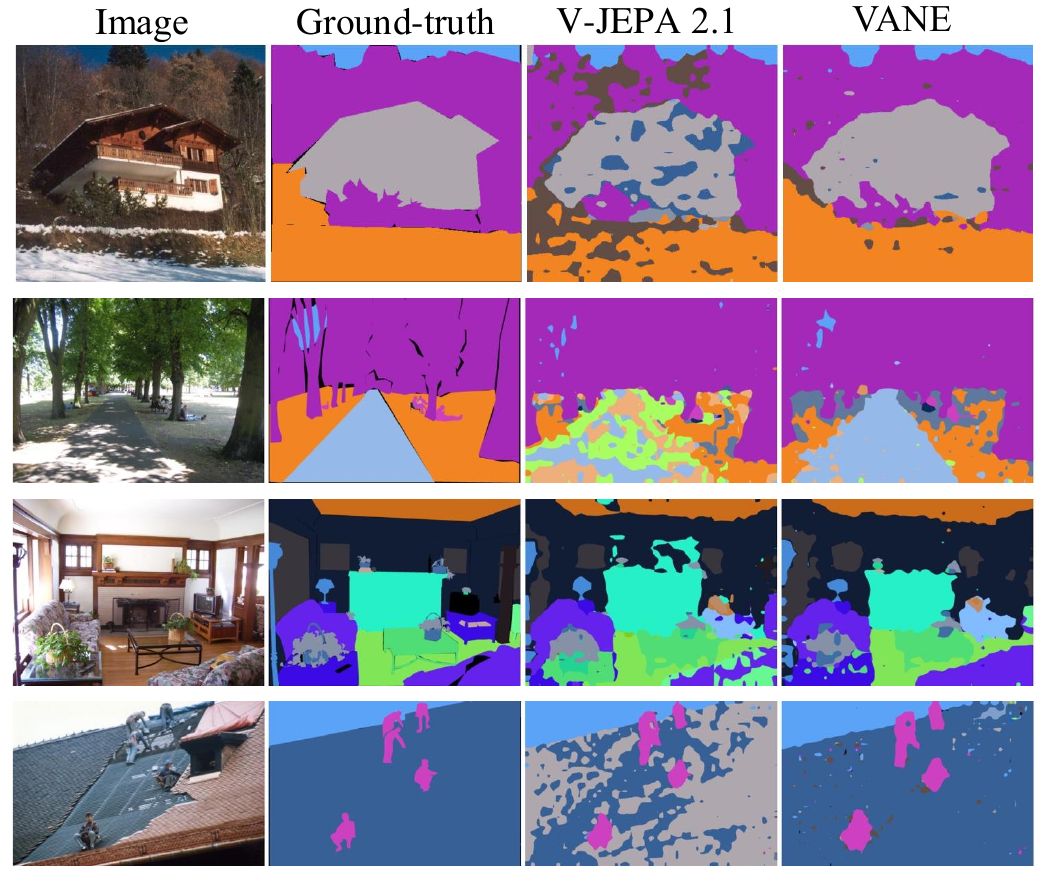}
\vspace{-14pt}
\caption{\textbf{Qualitative comparison on ADE20K.}
V-JEPA~2.1 and \method{} with the same dense linear probe on frozen
features.}
\label{fig:dense_feature_pca}
\vspace{-10pt}
\end{wrapfigure}
To assess spatial detail beyond global recognition, we evaluate semantic
segmentation on ADE20K, Cityscapes, and
VOC~\citep{zhou2017ade20k,cordts2016cityscapes,everingham2010voc} and
monocular depth estimation on NYUv2 and
KITTI~\citep{silberman2012nyuv2,geiger2012kitti}. Following
DINOv3~\citep{simeoni2025dinov3} and V-JEPA~2.1~\citep{mur2026vjepa21},
we train a dense linear projection on frozen final-layer patch features
for each task and report mIoU for segmentation and RMSE for depth.
Compared with V-JEPA~2.1 (Table~\ref{tab:dense_tasks}), \method{}
improves ADE20K and Cityscapes by 3.2 and 3.6 mIoU, respectively, while
remaining comparable on VOC (84.5 vs.\ 84.7).
% , while maintaining comparable performance on VOC (84.5 versus 84.7)
It also reduces depth RMSE
from 0.350 to 0.329 on NYUv2 and from 2.601 to 2.431 on KITTI, the lowest
errors among all encoders in Table~\ref{tab:dense_tasks}, including image
encoders. Since these benchmarks consist of still images, we attribute
the gains mainly to next-crop prediction and the generative decoder,
both of which improve ADE20K in our ablations
(Table~\ref{tab:ablations}b,c). Figure~\ref{fig:dense_feature_pca}
further shows that \method{} produces cleaner and more coherent segments
than V-JEPA~2.1.

\begin{table*}[t]
    \centering
    \vspace{-4mm}
    \small
    \caption{\textbf{Dense perception.}
Semantic segmentation (mIoU, \%) and monocular depth estimation (RMSE,
lower is better) with a linear probe on frozen final-layer patch
features. Param./patch denotes encoder size/patch size. Bold denotes the
best result among image--video encoders.}
    \label{tab:dense_tasks}
    \resizebox{\textwidth}{!}{%
    \begin{tabular}{lcccccc}
        \toprule
        \textbf{Method}
        & \textbf{Param./patch}
        & \multicolumn{3}{c}{\textbf{Segmentation}}
        & \multicolumn{2}{c}{\textbf{Depth}} \\
        \cmidrule(lr){3-5}\cmidrule(lr){6-7}
        & & ADE20K & Citysc. & VOC
        & NYUv2$\downarrow$ & KITTI$\downarrow$ \\
        \midrule
        \multicolumn{7}{l}{\textit{Image encoders evaluated under the same protocol}} \\
        % Web-DINO~\citep{fan2025webssl} & 7B/14 & 42.7 & 68.3 & 76.1 & 0.466 & 3.158 \\
        % DINOv3~\citep{simeoni2025dinov3} & 7B/16 & 55.9 & 81.1 & 86.6 & 0.309 & 2.346 \\
        % \midrule
        $\mathrm{PE}_{\mathrm{core}}$~\citep{bolya2025perceptionencoder} & 2B/14 & 38.9 & 61.1 & 69.2 & 0.590 & 4.119 \\
        $\mathrm{PE}_{\mathrm{spatial}}$~\citep{bolya2025perceptionencoder} & 2B/14 & 49.3 & 73.2 & 82.7 & 0.362 & 3.082 \\
        AM-RADIOv2.5~\citep{ranzinger2024radio} & 1B/14 & 53.0 & 78.4 & 85.4 & 0.340 & 2.918 \\
        SigLIP 2~\citep{tschannen2025siglip2} & 1B/16 & 42.7 & 64.8 & 72.7 & 0.494 & 3.273 \\
        DINOv2~\citep{oquab2023dinov2} & 1B/14 & 49.5 & 75.6 & 83.1 & 0.372 & 2.624 \\
        DINOv3 ViT-H+~\citep{simeoni2025dinov3} & 0.8B/16 & 54.8 & 79.5 & 85.8 & 0.352 & 2.635 \\
        \midrule
        \rowcolor{baselinebg}
        \multicolumn{7}{l}{\textit{Image--video encoders evaluated under the same protocol}} \\
        \rowcolor{baselinebg}
        InternVideo2-1B~\citep{wang2024internvideo2} & 1B/14 & 28.4 & 35.6 & 65.2 & 0.471 & 4.739 \\
        \rowcolor{baselinebg}
        V-JEPA 2~\citep{assran2025vjepa2} & 1B/16 & 24.4 & 45.9 & 63.9 & 0.642 & 4.650 \\
        \rowcolor{baselinebg}
        V-JEPA 2.1~\citep{mur2026vjepa21} & 1B/16 & 47.8 & 71.8 & \textbf{84.7} & 0.350 & 2.601 \\
        \rowcolor{napabg}
        % \method{} (ours) & 1B/16 & \textbf{50.6} & \textbf{74.3} & 84.5 & \textbf{0.345} & \textbf{2.580} \\
        \method{} (ours) & 1B/16 & \textbf{51.0} & \textbf{75.4} & 84.5 & \textbf{0.329} & \textbf{2.431} \\
        \bottomrule
    \end{tabular}}
    \vspace{-4mm}
\end{table*}

% Qualitative feature visualization remains pending in the supplied draft.
% Add its interpretation once the visualization and extraction settings are available.

\subsection{Ablation Studies}
\label{sec:ablation}
\label{sec:analysis}
\label{sec:scaling}

We ablate the key design choices of \method{}. Unless otherwise specified, all ablations use a ViT-L encoder trained for 50K iterations. Additional ablations on the predictor attention and the number of image crops, as well as a matched comparison with V-JEPA~2.1, are provided in Appendix~\ref{app:additional_ablation}. 
% A matched comparison with V-JEPA~2.1 is provided in Appendix~\ref{app:vista_controls}.
% We report IN1K, SSv2, and ADE20K to track their effects on appearance recognition, motion understanding, and spatial perception, respectively.

% \paragraph{Prediction space and granularity.}
% Replacing pixel targets with semantic embeddings from a frozen
% DINOv2 ViT-L tokenizer improves SSv2 by 16.2 points and ADE20K
% by 21.0 mIoU (Table~\ref{tab:ablation_target}), showing that
% prediction in semantic space transfers substantially better to motion
% recognition and dense perception. Among frozen tokenizers, DINOv2
% outperforms SigLIP~2 and VA-VAE by 6.8 and 4.3 points on SSv2,
% respectively. These comparisons suggest that effective prediction
% targets should combine visual discriminability with semantic
% abstraction. DINOv2 also improves SSv2 by 4.8 points over the online
% EMA target, indicating that a fixed pretrained target provides effective
% supervision for next-frame prediction. To isolate prediction granularity, we compare next-frame and next-token
% prediction using the same frozen DINOv2 ViT-L tokenizer. Next-frame
% prediction improves IN1K, SSv2, and ADE20K by 3.7, 19.1, and 22.7 points,
% respectively, with particularly large gains in motion and dense perception.
% This is consistent with Table~\ref{tab:global_tasks}, where \method{}
% outperforms the token-autoregressive Toto by 12.2 and 14.2 points on
% IN1K and K400. Details are provided in
% Appendix~\ref{app:prediction_granularity}.

\noindent\textbf{Prediction space and granularity.}
Replacing pixel targets with frozen DINOv2 ViT-L embeddings improves SSv2 by 16.2 points and ADE20K by 21.0 mIoU (Table~\ref{tab:ablations}a). 
Among alternative targets, DINOv2 outperforms SigLIP~2 and VA-VAE~\citep{yao2025reconstruction} by 6.8 and 4.3 points on SSv2, and the online EMA target by 4.8 points, suggesting that effective targets should capture the content and local structure of a scene while abstracting over appearance variation. We adopt DINOv2 as the default tokenizer because it performs best on IN1K and SSv2, whereas DINOv3 is only slightly better on ADE20K (45.2 vs.\ 45.1). 
To isolate the effect of prediction granularity, we also compare
next-frame prediction with standard next-token prediction, which
predicts each visual token from the preceding tokens
(Appendix~\ref{app:prediction_granularity}). With the target space
fixed, next-frame prediction improves IN1K, SSv2, and ADE20K by 3.7,
19.1, and 22.7 points, respectively, showing that the gains stem from
the prediction objective rather than from the target encoder itself.

% \begin{figure}[t]
% \centering

% % ==================== Table 6 ====================
% \begin{minipage}[t]{0.485\textwidth}
% \vspace{0pt}
% \centering
% \small
% \setlength{\tabcolsep}{2pt}
% \renewcommand{\arraystretch}{1.04}

% % Locally switch caption type to table.
% \makeatletter
% \def\@captype{table}
% \makeatother

% \caption{\textbf{Local-to-global image prediction.}
% Comparison of image prediction objectives with the video objective,
% tokenizer, and generative branch fixed. The default uses $K=3$.}
% \label{tab:ablation_state}

% \begin{tabularx}{\linewidth}{@{}Ycccc@{}}
% \toprule
% Image objective & $K$ & IN1K & SSv2 & ADE20K \\
% \midrule
% Masked image & -- & 81.0 & 69.3 & 38.3 \\
% Next cell & 4 & 84.0 & 72.6 & 45.1 \\
% \midrule
% Next crop & 2 & 84.3 & 72.1 & 44.4 \\
% \rowcolor{napabg}
% Next crop (default) & 3 & 84.5 & 72.3 & 45.1 \\
% Next crop & 4 & 84.4 & 71.9 & 44.7 \\
% \bottomrule
% \end{tabularx}

% \end{minipage}%
% \hfill
% % ==================== Figure 4 ====================
% \begin{minipage}[t]{0.485\textwidth}
% \vspace{0pt}
% \centering
% \setlength{\abovecaptionskip}{1pt}
% \setlength{\belowcaptionskip}{0pt}
% \includegraphics[width=\linewidth]{figures/recipe_progression.pdf}

% % Inherit figure type from the outer figure environment.
% \caption{\textbf{Cumulative training recipe.}}
% \label{fig:recipe_progression}

% \end{minipage}

% \end{figure}

\begin{table}[t]
\centering
\small
\vspace{4mm}
\caption{\textbf{Ablation studies.} Top-1 accuracy on IN1K/SSv2 and mIoU
on ADE20K with a ViT-L encoder trained for 50K iterations.
(a)~Prediction space and granularity; next-frame and next-token variants
share the frozen DINOv2 ViT-L target space. (b)~Image objectives, with
the video objective, tokenizer, and generative decoder fixed.
(c)~Auxiliary generative decoder.}
\label{tab:ablations}
\label{tab:ablation_target}
\setlength{\tabcolsep}{2pt}
\renewcommand{\arraystretch}{1.04}
\begin{minipage}[t]{0.52\textwidth}
\vspace{0pt}
\centering
\textbf{(a)} Prediction space and granularity\\[2pt]
\begin{tabularx}{\linewidth}{@{}Ylccc@{}}
\toprule
Target & Prediction & IN1K & SSv2 & ADE20K \\
\midrule
\multicolumn{5}{l}{\textit{Reference}} \\
Online EMA & Next frame & 81.0 & 67.5 & 35.3 \\
Pixels & Next frame & 75.8 & 56.1 & 24.1 \\
\midrule
\multicolumn{5}{l}{\textit{Semantic tokenizers}} \\
VA-VAE & Next frame & 81.8 & 68.0 & 37.1 \\
SigLIP 2 & Next frame & 77.5 & 65.5 & 29.6 \\
\rowcolor{napabg} DINOv2 ViT-L & Next frame & \textbf{84.5} & \textbf{72.3} & 45.1 \\
DINOv2 ViT-L & Next token & 80.8 & 53.2 & 22.4 \\
DINOv3 ViT-L & Next frame & 84.0 & 71.8 & \textbf{45.2} \\
\addlinespace[2pt]
\bottomrule
\end{tabularx}
\end{minipage}%
\hfill
\begin{minipage}[t]{0.44\textwidth}
\vspace{0pt}
\centering
\textbf{(b)} Image objectives\\[2pt]
\begin{tabularx}{\linewidth}{@{}Yccc@{}}
\toprule
Image objective & IN1K & SSv2 & ADE20K \\
\midrule
Masked image  & 81.0 & 69.3 & 38.3 \\
Next cell  & 83.7 & 71.6 & 43.1 \\
% \midrule
% Next crop & 2 & 84.3 & 72.1 & 44.4 \\
\rowcolor{napabg}
Next crop (default) & 84.5 & 72.3 & 45.1 \\
% Next crop & 4 & 84.4 & 71.9 & 44.7 \\
\bottomrule
\end{tabularx}

\vspace{6pt}
\textbf{(c)} Generative decoder\\[2pt]
\setlength{\tabcolsep}{4pt}
\begin{tabularx}{\linewidth}{@{}Yccc@{}}
\toprule
Setting & IN1K & SSv2 & ADE20K \\
\midrule
w/o Decoder
& 84.3
& 72.0
& 42.8 \\
\rowcolor{napabg} w/ Decoder
& \textbf{84.5}
& \textbf{72.3}
& \textbf{45.1} \\
\bottomrule
\end{tabularx}
\end{minipage}
\vspace{-4mm}
\end{table}

\noindent\textbf{Unified image prediction.} Table~\ref{tab:ablations}b compares three image objectives under the same video objective: masked-image completion, next-cell prediction, and local-to-global next-crop prediction. For next-cell prediction, we uniformly partition an image into a $2\times2$ grid, treat the four cells as an ordered sequence, and predict each cell from the preceding ones; next-crop prediction instead orders views from local to global and supervises only newly revealed regions. Next-crop prediction performs best on all three benchmarks, improving ADE20K by 2.0 and 6.8 mIoU over next-cell prediction and masked-image completion, respectively. Both ordered objectives outperform masked completion, indicating that images benefit from the same predictive structure as videos, and the further gain of next-crop prediction suggests that progressively expanding context is preferable to a fixed grid. 
% We use $T_i=3$ crops by default; 
% Appendix~\ref{app:crop_number} ablates the number of crops.

% \setlength{\intextsep}{1pt}

% \begin{wraptable}{r}{0.43\textwidth}
%     \centering
%     \vspace{-6pt}
%     \small
%     \caption{\textbf{Ablation of generative decoder.}}
%     \label{tab:gen_decoder_ablation}
%     \vspace{2pt}
%     \begin{tabular}{lccc}
%         \toprule
%         \textbf{Setting}
%         & \textbf{IN1K}
%         & \textbf{SSv2}
%         & \textbf{ADE20K} \\
%         \midrule
%         w/o Decoder
%         & 84.3
%         & 72.0
%         & 42.8 \\
%         w/ Decoder
%         & \textbf{84.5}
%         & \textbf{72.3}
%         & \textbf{45.1} \\
%         \bottomrule
%     \end{tabular}
%     \vspace{-4pt}
% \end{wraptable}

\noindent\textbf{Generative supervision for spatial detail.}
We test whether the auxiliary generative objective helps preserve spatial information beyond semantic prediction. Adding the generative decoder raises ADE20K from 42.8 to 45.1 mIoU (Table~\ref{tab:ablations}c), with smaller gains of 0.2 points on IN1K and 0.3 on SSv2. The larger gain on dense perception suggests reconstructing visual content in pixel space complements semantic prediction by preserving fine-grained spatial structure.

% \paragraph{Scaling and the training recipe.}
% Figure~\ref{fig:recipe_progression} examines whether the framework benefits
% from increased capacity, data, and temporal context. Scaling the encoder
% from 300M to 1B parameters improves IN1K by 1.4 points, SSv2 by 0.3,
% and ADE20K by 2.8 mIoU. Expanding the corpus further improves all three
% tasks, including a 1.2-point gain on SSv2. Extending clips from 16 to
% 64 frames for 60k additional iterations then raises SSv2 from 73.8 to
% 75.6, while also improving IN1K and ADE20K. Across the completed recipe,
% gains over the baseline reach 3.0 points on IN1K, 3.7 on SSv2, and
% 5.9 mIoU on ADE20K. These stages build cumulatively and include additional
% training compute. 
% The final 12k-iteration high-resolution cooldown increases the input resolution to $384\times384$, further improving motion recognition and dense perception.

% \begin{wrapfigure}{r}{0.42\textwidth}
% \vspace{-12pt}
% \centering
% \includegraphics[width=\linewidth]{figures/recipe_progression.pdf}
% \caption{\textbf{Cumulative training recipe.} Starting from the default ablation setting, each stage adds model scaling (ViT-L to ViT-g), data expansion, long-video post-training, or high-resolution cooldown to the previous configuration.}
% \label{fig:recipe_progression}
% \vspace{-10pt}
% \end{wrapfigure}
\begin{wrapfigure}{r}{0.45\textwidth}
\vspace{-12pt}
\centering
\includegraphics[width=\linewidth]{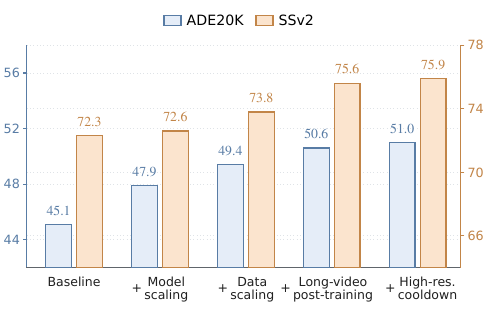}
\caption{\textbf{Cumulative training recipe.} 
% SSv2 top-1 accuracy and ADE20K mIoU after each cumulative stage, starting from the default ablation setting.
}
\label{fig:recipe_progression}
\vspace{-10pt}
\end{wrapfigure}
\noindent\textbf{Scaling and the training recipe.}
Figure~\ref{fig:recipe_progression} summarizes the cumulative effects of model scaling, data expansion, and post-training. 
Scaling the encoder from 300M to 1B parameters improves both SSv2 and
ADE20K, while data expansion brings further gains. 
Long-video post-training yields the clearest improvement on SSv2 (73.8 to 75.6), suggesting that longer temporal context strengthens next-frame prediction by exposing the model to richer visual evolution.
The final $384\times384$ cooldown further improves SSv2 and ADE20K. 
Overall, the complete recipe improves SSv2 and ADE20K by 3.6 and 5.9 points, respectively, highlighting complementary benefits from model and data scaling, longer temporal context, and higher spatial resolution.

% =============================================================================
% 5. CONCLUSION
% =============================================================================
\vspace{-1ex}
\section{Conclusion}
\label{sec:conclusion}
\vspace{-1ex}

We presented \method{}, a self-supervised visual pretraining framework
that predicts the next frame in embedding space rather than recovering
missing content within an observed context, making temporal evolution the
primary learning signal. Extending this objective to images through
local-to-global next-crop prediction enables unified image--video
pretraining.
% Despite using a frame-causal backbone, \method{} improves over the bidirectional V-JEPA~2.1 on anticipation (+7.1 recall@5 on EK100) and global recognition, and on most dense perception benchmarks. Together with gains from model and data scaling and multi-stage post-training, these results support next-frame prediction in semantic embedding space as a scalable principle for visual world modeling that captures both what the visual world contains and how it evolves.
Across anticipation, global recognition, and dense
perception, \method{} achieves state-of-the-art performance among unified
encoders on most benchmarks and improves consistently with model and data
scale, supporting next-frame prediction in embedding space as a scalable
principle for visual world modeling that captures both what the visual
world contains and how it evolves.

\clearpage

\subsection*{AI use statement}
Generative AI tools were used solely to assist with information search,
language editing, and polishing of the manuscript. All AI-assisted
content was reviewed and revised by the authors, who take full
responsibility for the final content of the paper.

\subsection*{Reproducibility statement}
We provide the details required to reproduce our method and experiments
throughout the main paper and appendix. The main paper specifies the
learning objectives, model components, training procedure, and downstream
evaluation protocols. The appendix provides additional details on dataset
construction and sampling, model architecture, video context masking,
stage-wise training configurations and hyperparameters, and evaluation
settings. For standard downstream benchmarks, we follow the established
protocols cited in the corresponding experimental sections.

\bibliographystyle{plainnat}
\bibliography{references}

\clearpage
% =============================================================================
% APPENDIX
% =============================================================================
\appendix

\newcommand{\tbd}{\textit{TBD}}
\section*{\Large{Appendix}}

% =============================================================================
% B. MODEL ARCHITECTURE
% =============================================================================
\section{Model Architecture}
\label{app:architecture}

VANE consists of a shared causal visual encoder, a context-isolated
state predictor, a frozen semantic tokenizer, and a lightweight
generative decoder. The encoder is the only component retained for
downstream evaluation; the state predictor, semantic tokenizer, and
generative decoder are used only during pretraining.

Table~\ref{tab:model_architecture} summarizes the model configurations.
Unless otherwise specified, the predictor uses the same depth and hidden dimension across encoder scales. 

\begin{table}[!htbp]
    \centering
    \small
    \setlength{\tabcolsep}{7pt}
    \caption{\textbf{Model architectures.}
    The main model uses a ViT-g causal encoder, a 12-layer
    context-isolated state predictor, a frozen DINOv2 ViT-L semantic
    tokenizer, and a six-layer patch-isolated DiffMLP decoder.}
    \label{tab:model_architecture}

    \begin{tabular}{lcc}
        \toprule
        & \textbf{ViT-L} & \textbf{ViT-g} \\
        \midrule

        \multicolumn{3}{l}{\textit{Causal Encoder}} \\

        Block
        & \multicolumn{2}{c}{ViT Block} \\

        Depth
        & 24 & 40 \\

        Width
        & 1024 & 1408 \\

        Num. Heads
        & 16 & 22 \\

        MLP Ratio
        & 4.0 & 4.36 \\

        Parameters
        & 300M & 1B \\

        Patch Size
        & \multicolumn{2}{c}{$16\times16$} \\

        Video Tubelet Size
        & \multicolumn{2}{c}{2 frames} \\

        Normalization
        & \multicolumn{2}{c}{LayerNorm} \\

        Position Encoding
        & \multicolumn{2}{c}{3D RoPE} \\

        Attention Mask
        & \multicolumn{2}{c}{Frame-causal} \\

        \midrule

        \multicolumn{3}{l}{\textit{State Predictor}} \\

        Block
        & \multicolumn{2}{c}{Context-isolated attention block} \\

        Depth
        & \multicolumn{2}{c}{12} \\

        Width
        & \multicolumn{2}{c}{384} \\

        Num. Heads
        & \multicolumn{2}{c}{12} \\

        Input Projection
        & \multicolumn{2}{c}{Encoder width $\rightarrow 384$} \\

        Output Projection
        & \multicolumn{2}{c}{$384 \rightarrow$ Encoder width} \\

        Normalization
        & \multicolumn{2}{c}{LayerNorm} \\

        Position Encoding
        & \multicolumn{2}{c}{3D RoPE} \\

        Prediction-token Init.
        & \multicolumn{2}{c}{Learned token duplicated at target positions} \\

        Dropout / DropPath
        & \multicolumn{2}{c}{0.0 / 0.0} \\

        \midrule

        \multicolumn{3}{l}{\textit{Semantic Tokenizer}} \\

        Architecture
        & \multicolumn{2}{c}{DINOv2 ViT-L} \\

        Patch Size
        & \multicolumn{2}{c}{$14\times14$} \\

        Optimization
        & \multicolumn{2}{c}{Frozen throughout pretraining} \\

        Feature
        & \multicolumn{2}{c}{Final patch-token representation} \\

        Per-frame Feature Dim.
        & \multicolumn{2}{c}{1024} \\

        % Target Normalization
        % & \multicolumn{2}{c}{LayerNorm without affine parameters} \\
        % Spatial Alignment
        % & \multicolumn{2}{c}{Resize tokenizer input; no feature interpolation} \\

        % Target Grid
        % & \multicolumn{2}{c}{Matched to encoder patch grid} \\

        Target Packing
        & \multicolumn{2}{c}{Video: two-frame concatenation; Image: duplication} \\

        Target Dimension
        & \multicolumn{2}{c}{2048} \\

        Semantic Projection
        & $1024\rightarrow2048$
        & $1408\rightarrow2048$ \\

        \midrule

        \multicolumn{3}{l}{\textit{Patch-Isolated Generative Decoder}} \\

        Block
        & \multicolumn{2}{c}{DiffMLP} \\

        Depth
        & \multicolumn{2}{c}{6} \\

        Hidden Dimension
        & \multicolumn{2}{c}{1536} \\

        Image Pixel Packing
        & \multicolumn{2}{c}{Duplicate along temporal dimension} \\

        Video Pixel Packing
        & \multicolumn{2}{c}{Two-frame tubelet} \\

        Noisy Pixel Dimension
        & \multicolumn{2}{c}{1536} \\

        Condition Injection
        & \multicolumn{2}{c}{Concatenation} \\

        Input Projection
        & $2560\rightarrow1536$
        & $2944\rightarrow1536$ \\

        Time Embedding
        & \multicolumn{2}{c}{$256\rightarrow1536\rightarrow1536$} \\

        Activation
        & \multicolumn{2}{c}{SiLU} \\

        Conditioning
        & \multicolumn{2}{c}{Predicted state $s_{t,j}$ and flow time $\tau$} \\

        Noisy Input
        & \multicolumn{2}{c}{$\tilde{x}_{t,j}(\tau)\in\mathbb{R}^{1536}$} \\

        Output
        & \multicolumn{2}{c}{1536-d flow velocity $\hat{v}_{t,j}$} \\

        Patch Interaction
        & \multicolumn{2}{c}{None (patch-isolated)} \\

        \bottomrule
    \end{tabular}
\end{table}

\begin{figure}[!ht]
    \centering
    \includegraphics[width=0.75\linewidth]{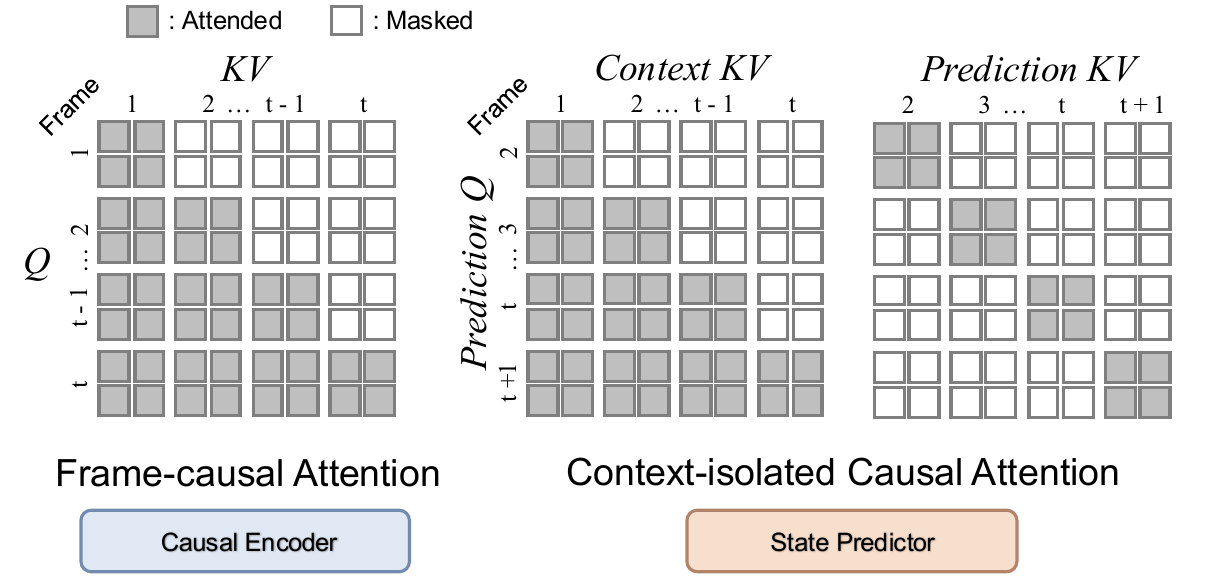}
    \caption{\textbf{Attention masks.} \emph{Left:} frame-causal attention
    in the causal encoder, bidirectional within each frame and
    unidirectional across frames. \emph{Right:} context-isolated causal
    attention in the state predictor. Prediction tokens for frame $t+1$ read
    the context states of frames up to $t$, which supply only keys and
    values, and attend to each other only within the same target frame.}
    \label{fig:attention_mask}
\end{figure}

\noindent\textbf{Causal encoder.}
The encoder processes all observations in a sequence in a single
forward pass. Under frame-causal attention, tokens in observation $t$
may attend to tokens from observations $1,\ldots,t$, but never to future
observations. Tokens within the same observation are processed jointly.
For videos, the input is patchified into $2\times16\times16$
temporal-spatial tubelets ($T\times H\times W$), such that each temporal
encoder position spans two consecutive sampled raw frames. For images,
each local-to-global crop constitutes one observation.
Images and videos use separate patch embeddings and a learned modality
embedding, while all other components are shared; image sequences use no
context masking.
Fig.~\ref{fig:attention_mask} (left) illustrates the attention mask.

\noindent\textbf{Context-isolated state predictor.}
The state predictor receives encoded historical states as read-only
context. Since the encoder and predictor use different hidden
dimensions, encoder states are first projected from the encoder width
to the 384-dimensional predictor space. Historical context states then
provide keys and values but are never updated inside the predictor.
Prediction tokens are obtained by duplicating a learned token and
assigning the corresponding target patch positions.

Prediction tokens for observation $t+1$ attend to the context states of
observations $1,\ldots,t$ and to one another, while predictions belonging
to different future observations remain isolated
(Fig.~\ref{fig:attention_mask}, right). The resulting predictor states are projected back to
the encoder width before being passed to the semantic and generative
objectives. This design separates representation learning from
future-state readout: the predictor can only use information already
encoded by the causal backbone.

\noindent\textbf{Semantic tokenizer.}
We use a frozen DINOv2 ViT-L as the semantic tokenizer for the main
model. We take its final patch tokens after removing special tokens and
normalize each patch feature with a LayerNorm without affine parameters;
no additional $\ell_2$ normalization is applied.

To align tokenizer targets with the encoder patch grid, we resize the
tokenizer input so that its $14\times14$ patchification produces the same
$H_p\times W_p$ spatial grid, without interpolating tokenizer features.
For example, a $256\times256$ encoder input produces a $16\times16$ grid,
while the corresponding tokenizer input is resized to $224\times224$.

For images, the tokenizer encodes the full image once, and crop targets
are indexed from the aligned full-image grid at their corresponding patch
positions. For videos, it encodes individual frames independently.
Each normalized patch feature is 1024-dimensional; features at the same
spatial position in a two-frame video tubelet are concatenated into a
2048-dimensional target, while image features are duplicated to the same
dimension. Both modalities therefore share the same semantic projection
head, which maps the predictor output to the 2048-dimensional tokenizer
target space.

\noindent\textbf{Patch-isolated generative decoder.}
The auxiliary generative decoder is implemented as a six-layer DiffMLP
with hidden dimension 1536. To share the same decoder across modalities,
each image is duplicated along the temporal dimension before pixel
patchification, forming a two-frame input with identical image content.
Consequently, both image patches and video tubelets are represented by
$2\times3\times16\times16=1536$ pixel values.

For the main ViT-g model, the 1408-dimensional predicted state is
concatenated with the 1536-dimensional noisy pixel token and projected
from 2944 to the 1536-dimensional decoder space. Flow time is encoded
using a 256-dimensional sinusoidal embedding followed by a
$256\rightarrow1536\rightarrow1536$ SiLU MLP. Each DiffMLP block contains
LayerNorm, an MLP, SiLU activations, and adaptive shift, scale, and gate
modulation. The final layer predicts a 1536-dimensional flow velocity.
Target patches are decoded independently, without access to neighboring
target patches or encoder context other than through the predicted state.

% =============================================================================
% A. PRETRAINING DATA
% =============================================================================

\section{Pretraining Data}
\label{app:data}

We jointly pretrain \method{} on large-scale image and video corpora.
Our corpus contains approximately 145M images and 24M video samples, matching V-JEPA~2.1~\citep{mur2026vjepa21} in overall image--video sample count. 
The data are constructed independently, following the large-scale curation principles of DINOv2~\citep{oquab2023dinov2} and V-JEPA~2.1~\citep{mur2026vjepa21}, and combine public datasets with additional retrieved and class-balanced data. 
For all public datasets, only their training splits are used for pretraining and retrieval-based curation; validation and test splits are excluded throughout data construction.

We refer to the resulting unified visual corpus as \textsc{VISTA}~(\textbf{V}isual \textbf{I}mage--video \textbf{S}calable \textbf{T}raining \textbf{A}rchive), with \textsc{VISTA-I} and \textsc{VISTA-V} denoting its image and video subsets, respectively.
Each subset combines public seed datasets with additional samples constructed through distribution-matched retrieval and class-balanced curation. 
The public datasets therefore serve both as high-quality training data and, where applicable, as reference distributions for expanding the corpus.

\subsection{Multi-Domain Data Curation}
\label{app:data_curation}

\noindent\textbf{Distribution-matched retrieval.}
Following DINOv2~\citep{oquab2023dinov2} and V-JEPA~2.1~\citep{mur2026vjepa21}, we use the training splits of multiple curated public datasets to define complementary visual distributions and retrieve visually matched samples from a substantially larger independently collected candidate pool. 
% Specifically, we extract visual
% representations for candidate samples and construct a large-scale
% nearest-neighbor index using \texttt{FAISS-GPU}. Samples from the
% reference datasets are then used as queries to retrieve visually
% related instances.

For images, the retrieval references span broad recognition, fine-grained recognition, textures, scenes, and object-centric domains, including IN21K, IN1K, Caltech-101, CUB-200-2011, DTD, FGVC-Aircraft, Flowers-102, Food-101, Oxford-IIIT Pet, Stanford Cars, SUN397, PASCAL VOC 2007, and WebVision.
Additional public image sources, including SA-1B, Google Landmarks v2, and MET, are incorporated directly into \textsc{VISTA-I}. 
Together, these sources provide broad coverage of objects, scenes, textures, fine-grained categories, and in-the-wild visual content.

% For videos, we analogously construct retrieval data from multiple temporal domains. SSv2 provides a motion-centric reference distribution, K710 covers diverse human actions, and COIN and EPIC-KITCHENS provide instructional and egocentric reference distributions. HowTo100M is additionally incorporated directly as large-scale instructional video data. The public video datasets and their retrieved counterparts are combined to form the distribution-matched component of \textsc{VISTA-V}.

For videos, we adopt the same retrieval reference datasets as
V-JEPA~2.1~\citep{mur2026vjepa21}, using the training splits of SSv2, K710, COIN, and EPIC-KITCHENS to cover motion-centric, action-centric, instructional, and egocentric domains, respectively.
HowTo100M is additionally incorporated directly as large-scale
instructional video data. The public video datasets and their
retrieved counterparts are combined to form the distribution-matched component of \textsc{VISTA-V}.

Importantly, the public datasets determine the reference distributions for retrieval, while the retrieved samples themselves are independently collected and are not drawn from the corresponding public datasets.
The public seed datasets are retained as part of \textsc{VISTA-I/V} rather than being replaced by their retrieved counterparts.

% TODO:
% Add the retrieval feature encoder, similarity metric, FAISS index type,
% top-k retrieval setting, and duplicate/filtering rules once finalized.

\noindent\textbf{Class-balanced curation.}
Distribution-matched retrieval preserves the visual characteristics of the reference datasets but can inherit their long-tailed semantic distributions. 
We therefore further augment both \textsc{VISTA-I} and \textsc{VISTA-V} with class-balanced data, following the data balancing principle of DINOv3~\citep{simeoni2025dinov3}. This complementary component broadens semantic coverage and reduces the dominance of frequent visual concepts in the retrieved data.

Overall, each \textsc{VISTA} subset combines three complementary components: public seed datasets, distribution-matched retrieved data, and class-balanced data. Applying the same curation principle to both modalities yields a unified image--video corpus for large-scale visual pretraining.

\subsection{Homogeneous and Heterogeneous Batch Sampling}
\label{app:data_sampling}

We combine homogeneous and heterogeneous sampling during pretraining, as summarized in Table~\ref{tab:pretrain_data}.
For $10\%$ of optimization steps, we construct homogeneous batches, sampling images exclusively from IN1K and videos from SSv2. These batches provide a stable learning signal from canonical image and video recognition datasets. 

For the remaining $90\%$ of steps, image and video batches are sampled from \textsc{VISTA-I} and \textsc{VISTA-V}, respectively. 
For video sampling, we organize \textsc{VISTA-V} into four source groups with sampling probabilities $0.20/0.20/0.25/0.35$, corresponding to motion-centric data, action-centric data, instructional and retrieved video data, and class-balanced video data, respectively. 
These probabilities are realized through proportional allocation of data-parallel ranks across the four source groups, with each rank sampling from its assigned group.

Image and video batches are processed in separate forward and backward passes, and their gradients are accumulated before each parameter update. 
This allows the two modalities to share the same encoder and predictive objective while retaining modality-specific data sampling.

\begin{table}[t]
    \centering
    \small
    \setlength{\tabcolsep}{10pt}
    \renewcommand{\arraystretch}{1.08}

    \caption{\textbf{Pretraining data mixture.}
    \textsc{VISTA-I} and \textsc{VISTA-V} denote the image and video
    subsets of our unified \textsc{VISTA} corpus. Each combines public
    seed datasets, distribution-matched retrieval, and class-balanced
    data as described in Sec.~\ref{app:data_curation}.}
    \label{tab:pretrain_data}

    \begin{tabular}{lll}
        \toprule
        \textbf{Regime}
        & \textbf{Modality}
        & \textbf{Training data} \\
        \midrule

        \multirow{2}{*}{\shortstack[l]{Homogeneous\\($10\%$)}}
        & Image & IN1K \\
        & Video & SSv2 \\

        \midrule

        \multirow{2}{*}{\shortstack[l]{Heterogeneous\\($90\%$)}}
        & Image & \textsc{VISTA-I} \\
        & Video & \textsc{VISTA-V} \\

        \bottomrule
    \end{tabular}
\end{table}

% =============================================================================
% C. TRAINING STRATEGY
% =============================================================================
\section{Multi-Stage Training Strategy}
\label{app:training}

Our training recipe contains four consecutive stages
(Table~\ref{tab:training_schedule}):
(1) learning-rate warm-up,
(2) large-scale base pretraining,
(3) long-video post-training, and
(4) high-resolution cooldown.
The first two stages learn the main representation using 16-frame clips
at $256\times256$ resolution, while the last two stages progressively
increase temporal and spatial context.

\subsection{Input Sampling and Augmentation}
\label{app:input_sampling}

\noindent\textbf{Video sampling.}
Videos are sampled at a target rate of 4 FPS. Base pretraining uses
16-frame clips and long-video post-training increases the observation
window to 64 frames. The encoder uses a temporal tubelet size of two,
so each temporal patch corresponds to two consecutive sampled frames.
For videos longer than the required observation window, a clip window
is randomly selected during training.

\noindent\textbf{Image and video augmentation.}
At $256\times256$ resolution, both image and video inputs use random
resized cropping with an area range of $[0.3,1.0]$ and aspect-ratio
range of $[0.75,1.35]$, followed by horizontal flipping with probability
0.5. We do not use AutoAugment, random erasing, or additional color
jitter in this training configuration. The same spatial augmentation
principle is retained when increasing the input resolution during
high-resolution cooldown.

\begin{figure}[!htbp]
    \centering
    \includegraphics[width=0.96\textwidth]
    {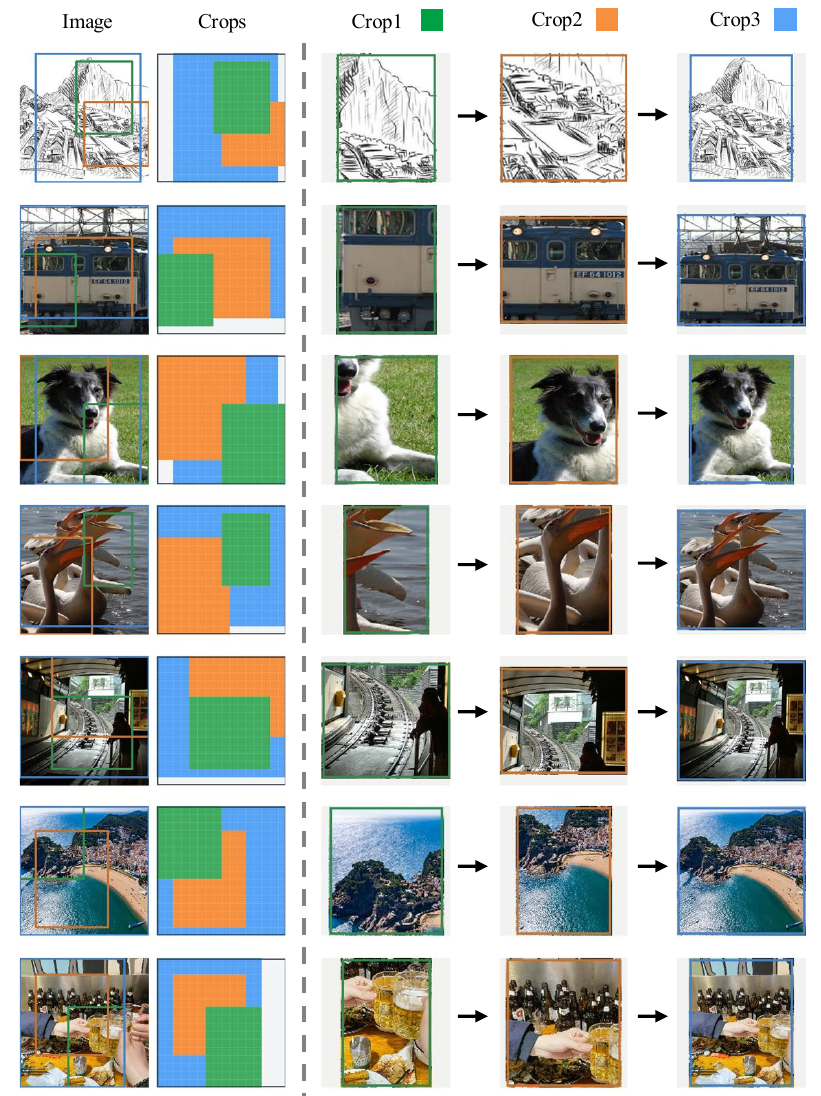}
    \caption{\textbf{Visualization of local-to-global next-crop prediction.} Each row shows the spatial layout of three sampled crops and the corresponding prediction sequence $\mathrm{Crop}_1 \rightarrow \mathrm{Crop}_2 \rightarrow \mathrm{Crop}_3$. The first two observations are local crops and the last provides broader scene context. At each step, supervision is restricted to spatial regions newly revealed by the next crop. Crops are resized to a common display size for visualization only; their original spatial coordinates are used during training.}
    \label{fig:next_crop_visualization}
\end{figure}

\begin{figure}[!htbp]
    \centering
    \includegraphics[width=0.96\textwidth]
    {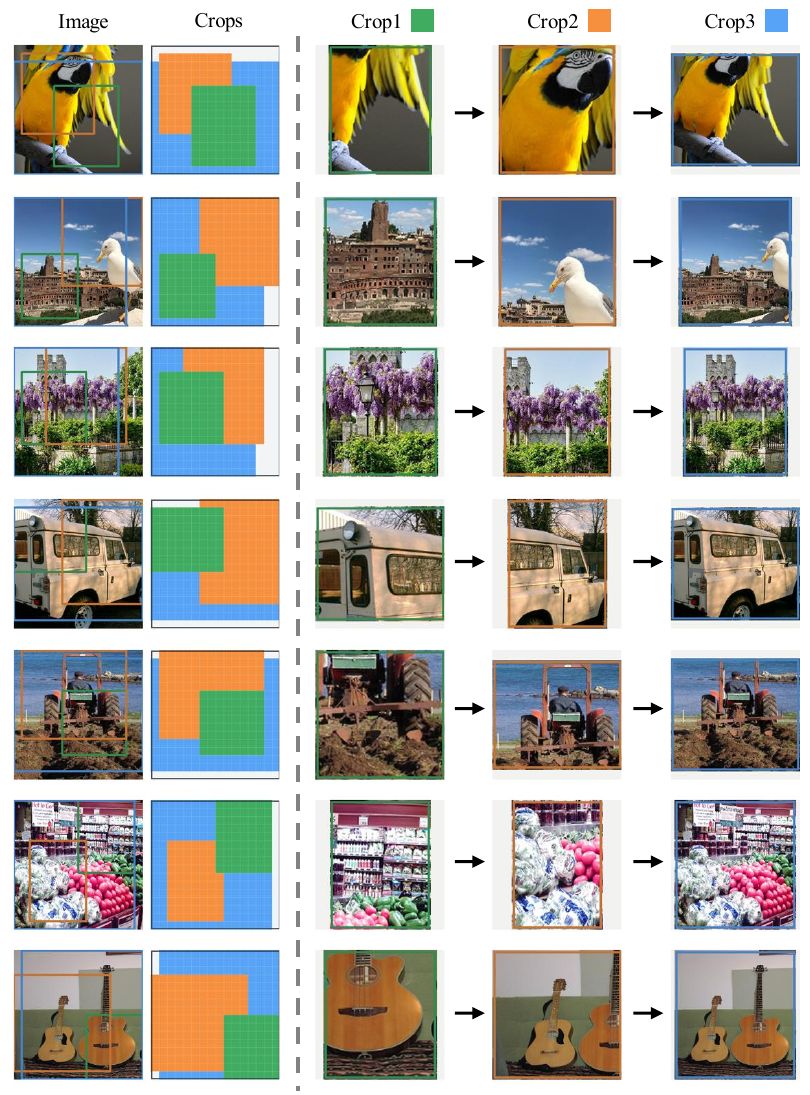}
    \caption{\textbf{Additional examples of local-to-global next-crop prediction.}
    We show additional sampled crop sequences following the same local-to-global
    sampling strategy as in Fig.~\ref{fig:next_crop_visualization}. Supervision is
    applied only to regions newly revealed by each subsequent crop. Crops are resized
    to a common display size for visualization only; their original spatial
    coordinates are used during training.}
    \label{fig:next_crop_visualization_more}
\end{figure}

\subsection{Local-to-Global Image Sampling}
\label{app:image_sampling}

We first apply the image augmentation described in Appendix~\ref{app:input_sampling} to obtain a base image.
Local-to-global views are then sampled from its patch grid without resizing the individual crops. All view coordinates and semantic targets are defined relative to this shared base image.

For image pretraining, we construct an ordered sequence of $K=3$ local-to-global crops. 
Two local crops are sampled with area ratios in $[0.2,0.6]$, followed by a global crop sampled with an area ratio in $[0.8,1.0]$. 
All crops use aspect ratios in $[0.75,1.35]$ and are sampled on the image patch grid.

The two local crops are ordered by area. Successive crops cover increasingly broad and complementary scene regions, and may partially overlap or extend to adjacent regions.
To encourage complementary observations, local-crop sampling favors pairs whose intersection occupies at most 0.4 of the larger local crop.
The final global crop provides substantially broader scene context but is not required to coincide exactly with the full uncropped image.
Semantic and generative supervision are applied only to patch positions newly revealed relative to preceding observations.

Figure~\ref{fig:next_crop_visualization} and Figure~\ref{fig:next_crop_visualization_more} visualize representative local-to-global crop sequences and their spatial relationships in the original image coordinate system.

\subsection{Video Context Masking}
\label{app:video_masking}

To increase the difficulty of next-frame prediction, we apply random multi-block tube masking to the historical video context while keeping future targets fully supervised, following V-JEPA~2.1~\citep{mur2026vjepa21}.

We follow the two multi-block mask configurations used in
V-JEPA~2.1. One samples eight spatial blocks with scale 0.15 while retaining at least 5\% of tokens; the other samples two larger blocks with spatial scale 0.70. 
For both configurations, the temporal scale is 1.0 and the block aspect ratio is sampled from $[0.75,1.5]$. 
Block positions are sampled randomly and may overlap, so the resulting fraction of masked context tokens varies across samples rather than being defined by a single fixed mask ratio. 
The spatial masking pattern is extended temporally to form tube masks across the observed clip.

\subsection{Warm-Up}
\label{app:warmup}

Training starts from randomly initialized VANE parameters while the
semantic tokenizer remains frozen. During the first 12K updates, the
learning rate is linearly increased from $1\times10^{-4}$ to
$6\times10^{-4}$. We use 16-frame video clips and $256\times256$
inputs, with global batch sizes of 1536 videos and 4608 images.
The warm-up stabilizes optimization before large-scale joint
image--video pretraining.

\subsection{Base Pretraining}
\label{app:base_pretrain}

After warm-up, we continue training for 180K updates using the same
16-frame, $256\times256$ input configuration and a constant learning
rate of $6\times10^{-4}$. This stage constitutes the main
representation-learning phase and uses the complete data mixture
described in Appendix~\ref{app:data}.

Image and video batches are processed independently in separate
forward/backward passes. Their gradients are accumulated before one
optimizer step, allowing the shared encoder and predictor to learn from
both modalities without requiring image and video samples to be packed
into the same sequence.

\begin{table}[!htbp]
    \centering
    \small
    \setlength{\tabcolsep}{5.0pt}
    \caption{\textbf{Multi-stage training strategy.}
    The base model is first trained with 16-frame clips at
    $256\times256$, followed by long-video post-training with 64-frame
    clips and a final high-resolution stage.}
    \label{tab:training_schedule}
    \begin{tabular}{lcccc}
        \toprule
        & \textbf{Warm-up}
        & \textbf{Base Pretrain}
        & \textbf{Long-Video}
        & \textbf{High-Res.} \\
        & & & \textbf{Post-train} & \textbf{Cooldown} \\
        \midrule

        \multicolumn{5}{l}{\textit{Input Configuration}} \\

        Video Frames
        & 16 & 16 & 64 & 64 \\

        Video Sampling Rate
        & \multicolumn{4}{c}{4 FPS} \\

        Video Tubelet Size
        & \multicolumn{4}{c}{2 frames} \\

        Resolution
        & $256^2$ & $256^2$ & $256^2$ & $384^2$ \\

        Image Views $K$
        & 3 & 3 & 3 & 3 \\

        Video Context Mask
        & \multicolumn{4}{c}{Random multi-block tube masking} \\

        \midrule

        \multicolumn{5}{l}{\textit{Optimization}} \\

        Updates
        & 12K & 180K & 60K & 12K \\

        Global Video Batch
        & 1536 & 1536 & 768 & 768 \\

        Global Image Batch
        & 4608 & 4608 & 4608 & 4608 \\

        Learning Rate
        & $1\times10^{-4}\rightarrow6\times10^{-4}$
        & $6\times10^{-4}$
        & $6\times10^{-4}$
        & $6\times10^{-4}\rightarrow6\times10^{-5}$ \\

        Optimizer
        & \multicolumn{4}{c}{AdamW} \\

        AdamW $\beta$
        & \multicolumn{4}{c}{$(0.9,\,0.95)$} \\

        AdamW $\epsilon$
        & \multicolumn{4}{c}{$10^{-8}$} \\

        Weight Decay
        & \multicolumn{4}{c}{0.04} \\

        Gradient Clipping
        & \multicolumn{4}{c}{1.0} \\

        Precision
        & \multicolumn{4}{c}{BF16} \\

        \midrule

        \multicolumn{5}{l}{\textit{Objectives}} \\

        Semantic Loss Weight
        & 1.0 & 1.0 & 1.0 & 1.0 \\

        $\lambda_{\mathrm{flow}}$
        & 0.25 & 0.25 & 0.25 & 0.25 \\

        Semantic Tokenizer
        & \multicolumn{4}{c}{Frozen DINOv2 ViT-L} \\

        \bottomrule
    \end{tabular}
\end{table}

\subsection{Long-Video Post-Training}
\label{app:long_video}

We next increase the temporal context from 16 to 64 frames while retaining the $256\times256$ spatial resolution. This stage runs for 60K additional updates with a global video batch size of 768, providing a longer observation window for predictive pretraining.

\subsection{High-Resolution Cooldown}
\label{app:highres}

The final stage increases the spatial input resolution from $256\times256$ to $384\times384$ for 12K updates while retaining the 64-frame observation window introduced during long-video post-training. It starts from the long-video checkpoint and linearly decays the learning rate during this stage.

The warm-up and base-pretraining stages together require approximately
26.6K GPU-hours, while the two post-training stages require
approximately 30.7K GPU-hours in total.

% =============================================================================
% D. ADDITIONAL RESULTS
% =============================================================================
\section{Additional Results}
\label{app:additional_results}

\subsection{Qualitative Short-Term Anticipation}
\label{app:ego4d_visualization}

\begin{figure}[!htbp]
    \centering
    \includegraphics[width=\textwidth]
    {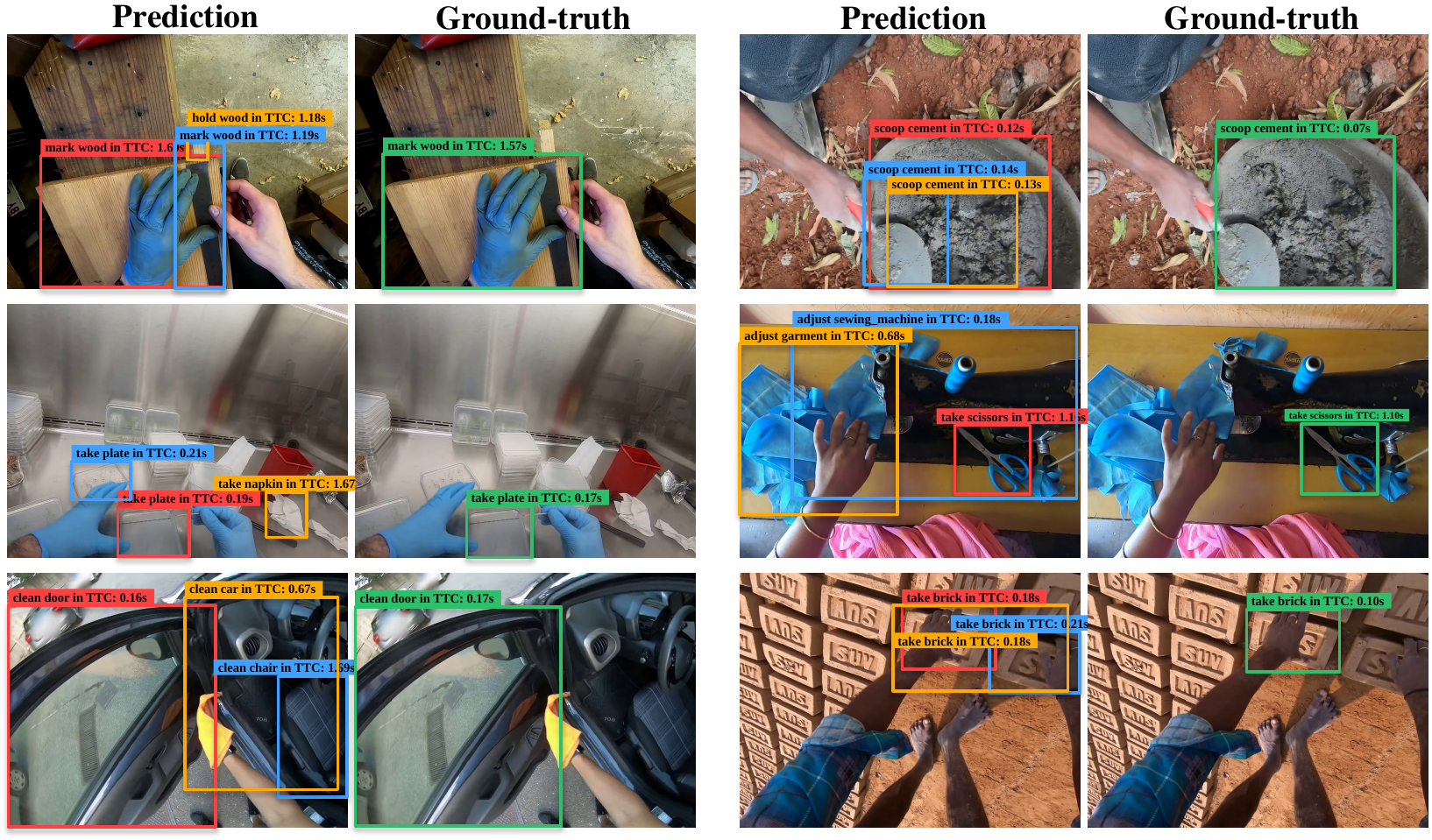}
    \caption{\textbf{Qualitative short-term anticipation results on Ego4D
    STA v2.} Each example pairs the top-scoring predictions of \method{}
    (left) with the ground truth (right). Each box marks a next-active
    object and is labeled with the interaction verb, the object noun, and
    the time to contact (TTC). Predictions are produced by a probe trained
    on frozen \method{} features.}
    \label{fig:ego4d_visualization}
\end{figure}

Figure~\ref{fig:ego4d_visualization} shows qualitative results on Ego4D
STA v2, obtained with the same frozen-feature probe as in
Table~\ref{tab:ego4d}. Each prediction consists of a bounding box for the
next-active object and a label combining the interaction verb, the object
noun, and the time to contact (TTC). In every example shown, one of the
top-scoring predictions matches the ground-truth verb--noun pair and
localizes the correct object with a comparable time to contact. The
remaining candidates are typically plausible alternatives in the same
scene, such as holding rather than marking the wood, or adjusting the
garment rather than taking the scissors.

\subsection{Qualitative Dense Perception}
\label{app:dense_visualization}

We provide additional qualitative results for semantic segmentation and
monocular depth estimation. All predictions are produced using the same
frozen-feature evaluation protocol as in the quantitative experiments
in the main paper. These examples complement aggregate metrics by
showing how well pretrained features preserve object boundaries,
fine-grained scene structure, and geometric layout.

\begin{figure}[!htbp]
    \centering
    \includegraphics[width=0.78\textwidth]
    {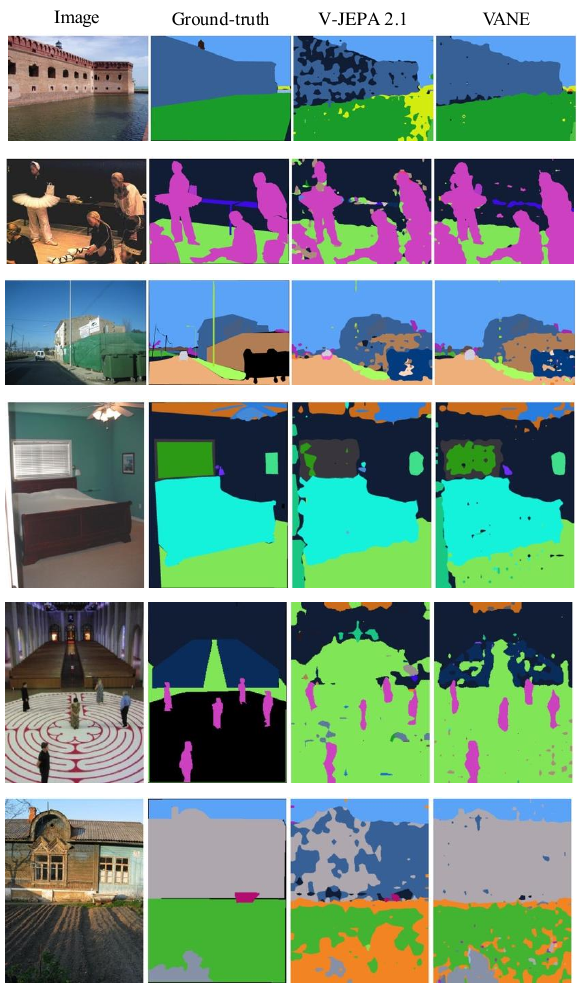}
    \caption{\textbf{Qualitative semantic segmentation results on ADE20K.}
    Each example shows the input image, ground truth, and predictions
    from dense probes trained on frozen features. All compared encoders
    use the same downstream protocol.}
    \label{fig:ade20k_visualization}
\end{figure}

\begin{figure}[!htbp]
    \centering
    \includegraphics[width=1.0\textwidth]
    {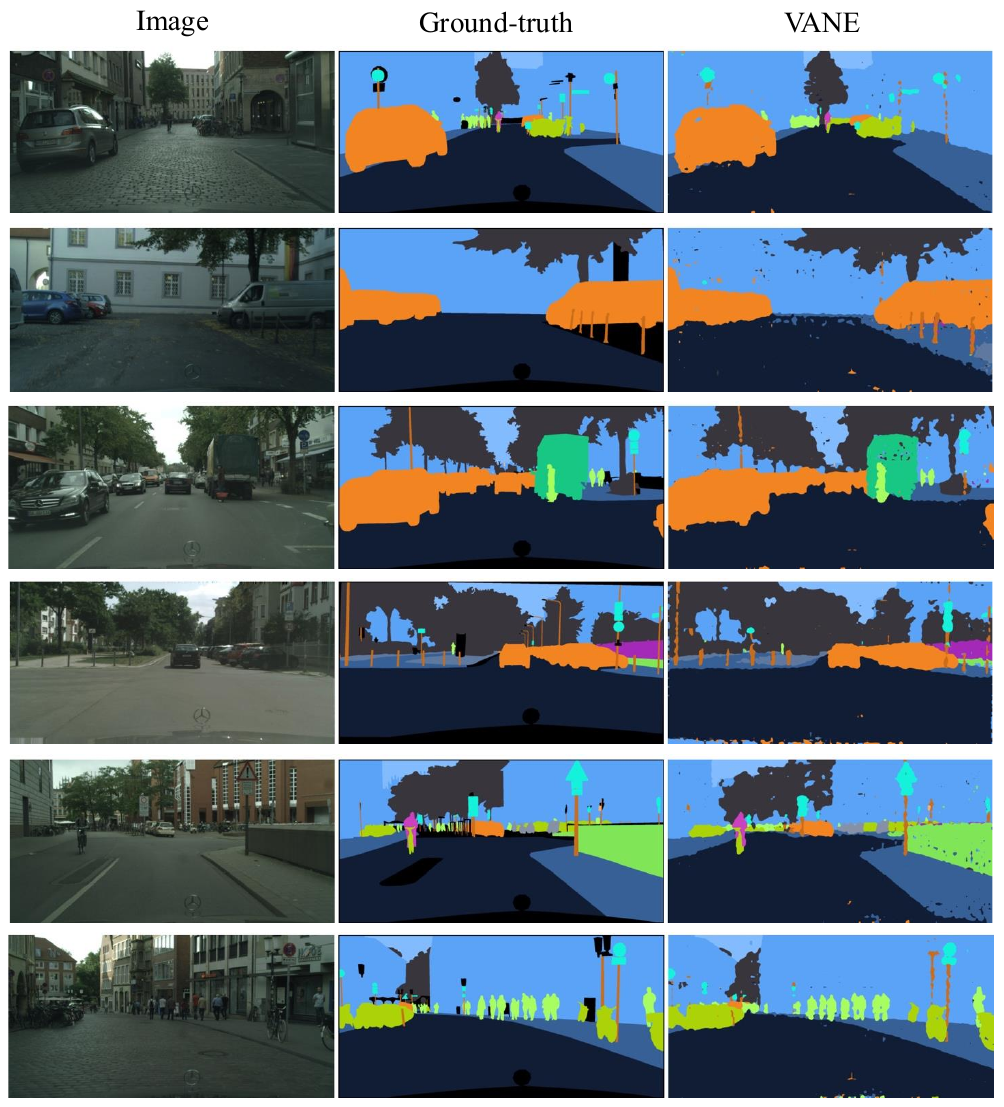}
    \caption{\textbf{Qualitative semantic segmentation results on Cityscapes.} Each example shows the input image, ground truth, and the prediction from a dense linear probe trained on frozen \method{} features.}
    \label{fig:cityscapes_visualization}
\end{figure}

\noindent\textbf{Semantic segmentation.}
Figures~\ref{fig:ade20k_visualization} and
\ref{fig:cityscapes_visualization} show representative examples from
ADE20K and Cityscapes. We visualize the input image, ground-truth
annotation, and predictions obtained from identical dense readouts
trained on frozen visual features.

\begin{figure}[!htbp]
    \centering
    \includegraphics[width=1.0\textwidth]
    {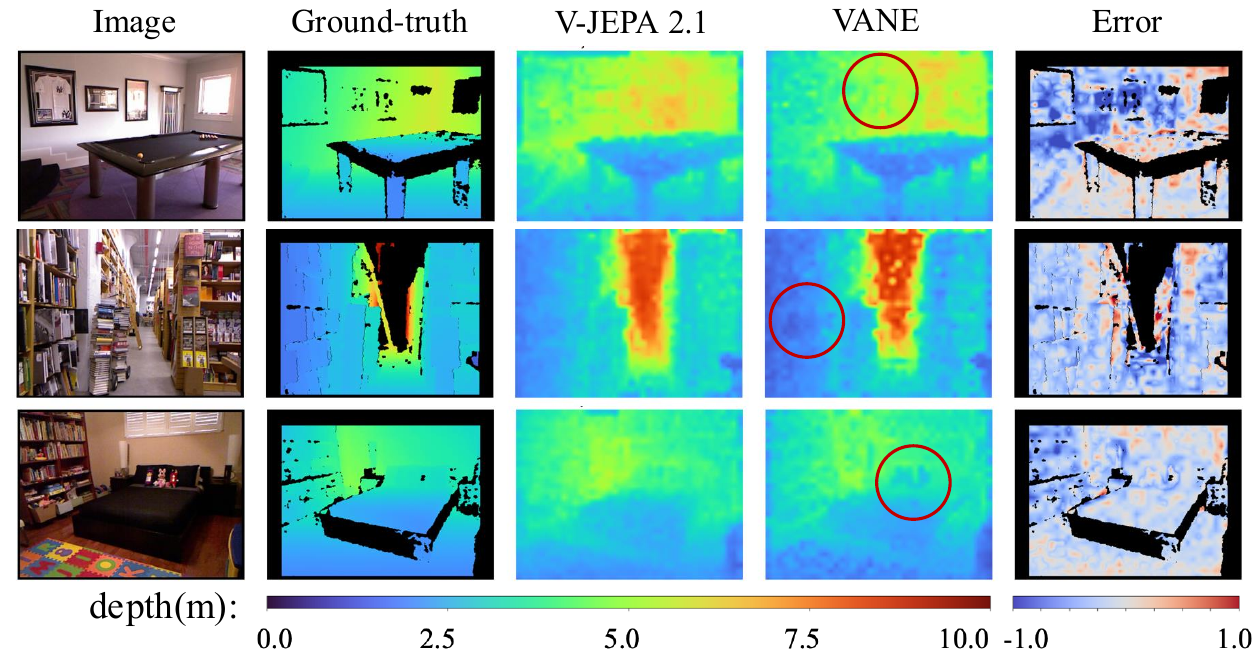}
    \caption{\textbf{Qualitative monocular depth estimation on NYUv2.} The error column shows $\Delta e= \left|\hat d_{\method{}}-d_{\mathrm{gt}}\right| -\left|\hat d_{\text{V-JEPA 2.1}}-d_{\mathrm{gt}}\right|$; blue favors \method{}, while red favors V-JEPA~2.1.}
    \label{fig:nyuv2_visualization}
\end{figure}

\begin{figure}[!htbp]
    \centering
    \includegraphics[width=1.0\textwidth]
    {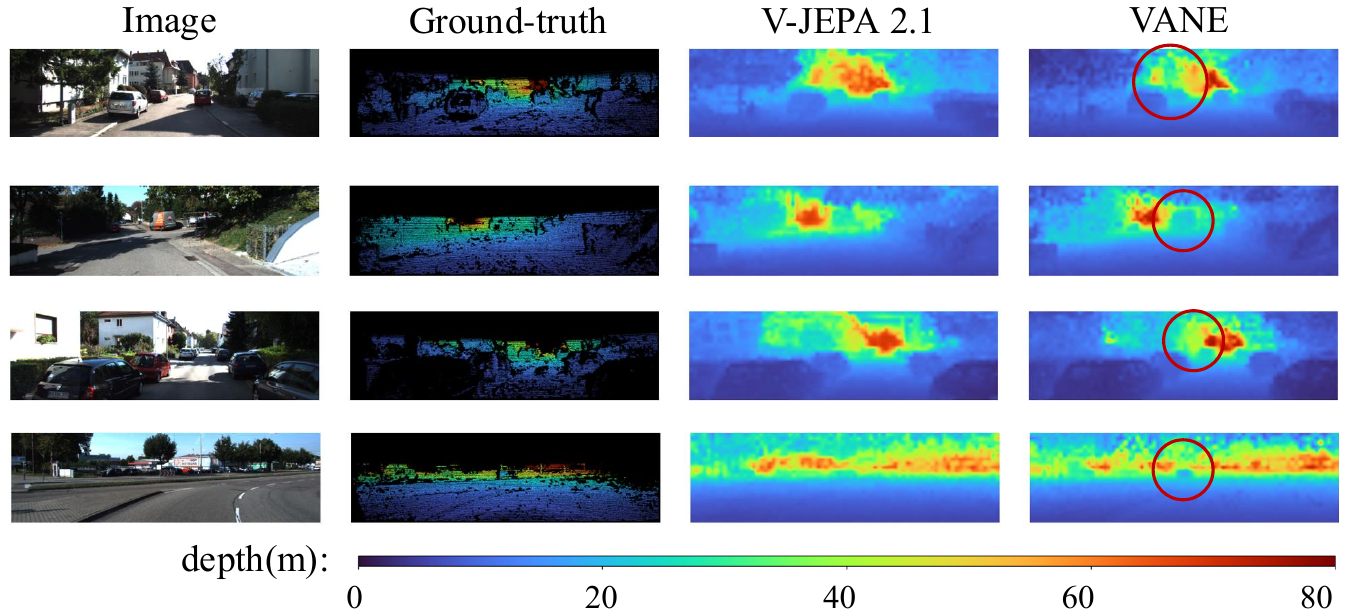}
    \caption{\textbf{Qualitative monocular depth estimation results on KITTI.}
    Predicted depth maps are obtained from dense readouts trained on
    frozen visual features using the same protocol as the quantitative
    evaluation.}
    \label{fig:kitti_visualization}
\end{figure}

\noindent\textbf{Monocular depth estimation.}
Figures~\ref{fig:nyuv2_visualization} and
\ref{fig:kitti_visualization} provide corresponding qualitative results
on NYUv2 and KITTI. NYUv2 emphasizes indoor geometric structure,
whereas KITTI provides long-range outdoor scenes and substantially
different depth statistics.

% =============================================================================
% E. ADDITIONAL ABLATIONS
% =============================================================================

\section{Additional Ablation Studies}
\label{app:additional_ablation}

The main paper studies prediction targets, image prediction strategies, generative supervision, scaling, and the training schedule. Here we first provide a controlled comparison with V-JEPA~2.1 under matched pretraining conditions, followed by further analyses of image crops, state-predictor attention, and prediction granularity. Unless otherwise specified, all ablations use a ViT-L encoder trained for 50K iterations.

\subsection{Masked versus Next-Frame Prediction under Matched Pretraining}
\label{app:vista_controls}

We perform a controlled comparison with V-JEPA~2.1~\citep{mur2026vjepa21}
to examine whether the gains of \method{} persist after accounting for
differences in pretraining data and prediction targets. All models use
ViT-L encoders trained on \textsc{VISTA} (Appendix~\ref{app:data}) for
50K iterations under the same training budget. V-JEPA~2.1 retains its
bidirectional masked feature-prediction formulation, whereas \method{}
uses next-frame prediction. We evaluate frozen encoder features using
the same downstream protocols as in the main paper, and report top-1
accuracy on IN1K and SSv2 and mIoU on ADE20K.

\begin{table}[t]
    \centering
    \small
    \setlength{\tabcolsep}{3.5pt}
    \renewcommand{\arraystretch}{1.08}
    \caption{\textbf{Masked versus next-frame prediction under matched
    pretraining.}
    All models use ViT-L encoders trained on \textsc{VISTA} for 50K
    iterations. The frozen-target V-JEPA~2.1 variant uses the same
    DINOv2 ViT-L semantic tokenizer as \method{}. We report top-1
    accuracy (\%) on IN1K and SSv2 and mIoU (\%) on ADE20K.}
    \label{tab:vista_target_control}

    \begin{tabular}{lllccc}
        \toprule
        \textbf{Method}
        & \textbf{Objective}
        & \textbf{Target}
        & \textbf{IN1K}
        & \textbf{SSv2}
        & \textbf{ADE20K} \\
        \midrule

        V-JEPA~2.1
        & Masked
        & EMA
        & 81.8 & 72.0 & 38.4 \\

        V-JEPA~2.1
        & Masked
        & Frozen DINOv2
        & 82.7 & 70.6 & 41.5 \\

        \midrule

        \method{}
        & Next frame
        & Frozen DINOv2
        & \textbf{84.5}
        & \textbf{72.3}
        & \textbf{45.1} \\

        \bottomrule
    \end{tabular}%
\end{table}

\noindent\textbf{Matching the pretraining data.}
We first compare \method{} with the standard V-JEPA~2.1 formulation
while matching the pretraining corpus, encoder scale, and iteration
budget. As shown in Table~\ref{tab:vista_target_control}, \method{}
improves IN1K accuracy by 2.7 points and ADE20K by 6.7 mIoU, while
achieving slightly higher SSv2 accuracy (72.3\% vs.\ 72.0\%). Thus, the
performance differences persist when both methods are trained on the
same data at the same model scale and training budget.

\noindent\textbf{Matching the semantic prediction space.}
We further replace the EMA target encoder of V-JEPA~2.1 with the same
frozen DINOv2 ViT-L~\citep{oquab2023dinov2} tokenizer used by
\method{}. This variant retains V-JEPA~2.1's bidirectional masked
feature-prediction objective while predicting targets in the same
semantic embedding space as \method{}.

The frozen DINOv2 target improves the masked baseline on IN1K and
ADE20K, from 81.8 to 82.7 and from 38.4 to 41.5, respectively, while
SSv2 decreases from 72.0 to 70.6. With both the pretraining corpus and
semantic target matched, \method{} retains gains of 1.8 points on IN1K,
1.7 points on SSv2, and 3.6 mIoU on ADE20K. Thus, neither the
pretraining corpus nor the frozen DINOv2 target alone explains the
performance gap between masked and next-frame pretraining.

\subsection{Number of Image Crops}
\label{app:crop_number}

Our default local-to-global image objective uses $K=3$ local-to-global crops. We additionally vary the number of crops to examine
how the amount of intermediate visual context affects representation
learning. For all settings, crops are ordered from local to global, and
supervision is restricted to regions newly revealed by the next
crop. The video objective, semantic tokenizer, model architecture, and
training schedule are kept fixed.

\begin{table}[t]
    \centering
    \small
    \setlength{\tabcolsep}{6pt}
    \caption{\textbf{Ablation of the number of image crops.}
    We vary the number of local-to-global views $K$ while keeping the
    remaining training configuration fixed.}
    \label{tab:crop_number_ablation}
    \begin{tabular}{cccc}
        \toprule
        \textbf{Number of Crops $K$}
        & \textbf{IN1K}
        & \textbf{SSv2}
        & \textbf{ADE20K} \\
        \midrule
        2
        & 84.3 & 72.1 & 44.4 \\

        % \rowcolor{napabg}
        3 (default)
        & \textbf{84.5} & \textbf{72.3} & \textbf{45.1} \\
        4
        & 84.4 & 71.9 & 44.7 \\
        \bottomrule
    \end{tabular}
\end{table}

As shown in Table~\ref{tab:crop_number_ablation}, performance is relatively stable across the three settings, with $K=3$ providing the best overall results. Using only two views provides less intermediate context during local-to-global prediction, while increasing to four views brings no further improvement. We therefore use $K=3$ as the default throughout the main experiments.

\subsection{State-Predictor Attention}
\label{app:predictor_attention}

We next study how the state predictor should interact with historical
encoder states. Our default predictor uses
\emph{context-isolated cross-attention}, where encoded historical states
are treated as read-only context and provide keys and values to the
predictor. Prediction tokens are updated by attending to this historical
context to produce the future visual state, while the encoded context
itself is not jointly updated inside the predictor.

As a control, we replace context-isolated attention with causal
self-attention that jointly updates historical states and prediction
tokens. Both variants enforce the same causal boundary: the prediction
for observation $t$ depends only on observations preceding $t$, and
prediction-token groups remain isolated across target observations.
The predictor depth, hidden dimension, target positions, and training
objective are otherwise unchanged.

\begin{table}[t]
    \centering
    \small
    \caption{\textbf{Ablation of state-predictor attention.}
    Context-isolated prediction treats historical encoder states as
    read-only context rather than jointly updating them with prediction
    tokens inside the predictor.}
    \label{tab:predictor_attention_ablation}
    \begin{tabular}{lccc}
        \toprule
        \textbf{Predictor Attention}
        & \textbf{IN1K}
        & \textbf{SSv2}
        & \textbf{ADE20K} \\
        \midrule
        Causal self-attention
        & 81.2 & 66.8 & 35.5 \\

        Context-isolated cross-attention
        & 84.5 & 72.3 & 45.1 \\
        \bottomrule
    \end{tabular}
\end{table}

Table~\ref{tab:predictor_attention_ablation} shows that
context-isolated prediction consistently improves downstream
representation quality. Compared with causal self-attention, it
improves IN1K accuracy by 3.3 points and SSv2 accuracy by 5.5 points,
with a larger gain of 9.6 mIoU on ADE20K. 

These results favor a separation between history encoding and future-state prediction. Causal self-attention allows the predictor to further transform historical states before producing the target representation. Context isolation instead requires the encoder outputs to carry the historical information needed for prediction, leaving the predictor to read these states and construct the future representation. This division of roles improves frozen-feature transfer in our experiments, particularly for dense perception.

\subsection{Prediction Granularity}
\label{app:prediction_granularity}

We provide implementation details for the next-token baseline used in the
prediction-granularity comparison in Table~\ref{tab:ablation_target}.
We compare our next-frame objective with standard next-token prediction
(NTP), while keeping the frozen DINOv2 ViT-L semantic target, encoder
scale, training schedule, batch size, input resolution, and clip length
fixed.

In standard NTP, visual patches are ordered first temporally and then
spatially. When predicting the $k$-th token of frame $t$, the model
observes all preceding frames and the first $k-1$ tokens of the current
frame, but not later tokens. The encoder therefore follows token-causal
attention, analogous to a conventional autoregressive Transformer.

Next-frame prediction instead treats the entire frame as the prediction
unit. All patches within an observed frame are jointly available, while
no patch from the target frame is observed. Accordingly, the encoder
uses frame-causal attention, allowing tokens to interact within each
observed frame and attend to preceding frames, but not future frames.
The target frame is then predicted jointly from the preceding
observations.

Both variants use the same semantic and generative objectives and differ
in how the visual sequence is factorized for prediction: NTP predicts
successive visual tokens from prior tokens, whereas next-frame prediction
predicts the next visual observation from preceding observations.

\end{document}